\def\isarxiv{1}

\ifdefined\isarxiv

\documentclass[11pt]{article}
\usepackage[authoryear]{natbib}

\else

\fi

\usepackage{amsmath}
\usepackage{amsthm}
\usepackage{amssymb}
\usepackage{amsfonts}
\usepackage{bbm}

\usepackage{algorithm}
\usepackage{algorithmicx}
\usepackage[noend]{algpseudocode}

\usepackage{graphicx}
\usepackage{grffile}
\usepackage{wrapfig,epsfig}
\usepackage{url}
\usepackage{epstopdf}
\usepackage{enumitem}

\usepackage{booktabs} 
\usepackage{multirow}
\usepackage{makecell}
\usepackage{threeparttable}
\usepackage{pgfplots}
\usepackage{pgfplotstable}
\pgfplotsset{compat=1.18}
\usepackage{subcaption}

\usepackage{dsfont}

\usepackage{xcolor}
\definecolor{BrickRed}{rgb}{0.8,0.25,0.33}
\definecolor{ForestGreen}{rgb}{0.1333,0.5451,0.1333}

\usepackage{hyperref}
\usepackage[capitalize,noabbrev,sort,nameinlink]{cleveref} 
\hypersetup{colorlinks={true},linkcolor={BrickRed},citecolor={ForestGreen}}

\allowdisplaybreaks
 
\ifdefined\isarxiv

\usepackage{tikz}
\usepackage{hyperref}
\usetikzlibrary{arrows}
\usepackage[margin=1in]{geometry}
\usepackage[T1]{fontenc}
\usepackage{microtype}
\usepackage{zlmtt}
\usepackage{pifont}
\usepackage[capitalize,noabbrev,sort,nameinlink]{cleveref} 

\else

\usepackage{microtype}

\fi
 
\graphicspath{{./figs/}}

\makeatletter
\newcommand*{\RN}[1]{\expandafter\@slowromancap\romannumeral #1@}
\makeatother

\usepackage{lineno}

\usepackage[many]{tcolorbox}
\theoremstyle{definition}
\newtheorem{theorem}{Theorem}[section]
\newtheorem{lemma}[theorem]{Lemma}
\newtheorem{definition}[theorem]{Definition}

\newtheorem{corollary}[theorem]{Corollary}

\newtheorem{assumption}[theorem]{Assumption}

\theoremstyle{plain}

\newtheorem{remark}[theorem]{Remark}

\tcolorboxenvironment{theorem}{
  breakable,
  colback=gray!20,
  colframe=white,
  boxrule=0pt,
  left=2pt,right=2pt,top=2pt,bottom=2pt,
  sharp corners,
  before skip=\topsep,
  after skip=\topsep
}
\tcolorboxenvironment{lemma}{
  breakable,
  colback=gray!20,
  colframe=white,
  boxrule=0pt,
  left=2pt,right=2pt,top=2pt,bottom=2pt,
  sharp corners,
  before skip=\topsep,
  after skip=\topsep
}
\tcolorboxenvironment{definition}{
  breakable,
  colback=gray!20,
  colframe=white,
  boxrule=0pt,
  left=2pt,right=2pt,top=2pt,bottom=2pt,
  sharp corners,
  before skip=\topsep,
  after skip=\topsep
}
\tcolorboxenvironment{corollary}{
  breakable,
  colback=gray!20,
  colframe=white,
  boxrule=0pt,
  left=2pt,right=2pt,top=2pt,bottom=2pt,
  sharp corners,
  before skip=\topsep,
  after skip=\topsep
}
\tcolorboxenvironment{assumption}{
  breakable,
  colback=gray!20,
  colframe=white,
  boxrule=0pt,
  left=2pt,right=2pt,top=2pt,bottom=2pt,
  sharp corners,
  before skip=\topsep,
  after skip=\topsep
}

\newcommand{\wh}{\widehat}
\newcommand{\wt}{\widetilde}

\newcommand{\N}{\mathcal{N}}
\newcommand{\R}{\mathbb{R}}

\renewcommand{\tilde}{\wt}
\renewcommand{\hat}{\wh}

\DeclareMathOperator*{\E}{{\mathbb{E}}}

\DeclareMathOperator*{\argmin}{\arg\min}

\DeclareMathOperator{\Diag}{Diag}

\DeclareMathOperator{\vect}{vec}
\DeclareMathOperator{\tr}{tr}

\newcommand{\bb}{\mathbf{b}}

\newcommand{\bv}{\mathbf{v}}
\newcommand{\bw}{\mathbf{w}}
\newcommand{\bx}{\mathbf{x}}
\newcommand{\by}{\mathbf{y}}
\newcommand{\bz}{\mathbf{z}}

\newcommand{\bA}{\mathbf{A}}
\newcommand{\bB}{\mathbf{B}}

\newcommand{\bD}{\mathbf{D}}
\newcommand{\bE}{\mathbf{E}}

\newcommand{\bH}{\mathbf{H}}
\newcommand{\bI}{\mathbf{I}}

\newcommand{\bM}{\mathbf{M}}

\newcommand{\bP}{\mathbf{P}}

\newcommand{\bW}{\mathbf{W}}
\newcommand{\bX}{\mathbf{X}}
\newcommand{\bY}{\mathbf{Y}}
\newcommand{\bZ}{\mathbf{Z}}

\newcommand{\cF}{\mathcal{F}}
\newcommand{\cG}{\mathcal{G}}
\newcommand{\cH}{\mathcal{H}}

\newcommand{\cL}{\mathcal{L}}

\newcommand{\cR}{\mathcal{R}}

\newcommand{\cT}{\mathcal{T}}

\newcommand{\cX}{\mathcal{X}}
\newcommand{\cY}{\mathcal{Y}}
\newcommand{\cZ}{\mathcal{Z}}

\newcommand{\bTheta}{\boldsymbol{\Theta}}
\newcommand{\bvareps}{\boldsymbol{\varepsilon}}
\newcommand{\bVareps}{\boldsymbol{\mathcal{E}}}

\begin{document}

\ifdefined\isarxiv

\date{}

\title{Implicit Hypergraph Neural Networks: A Stable Framework for Higher-Order Relational Learning with Provable Guarantees}
\author{
Xiaoyu Li\thanks{ \texttt{xiaoyu.li2@student.unsw.edu.au}. Equal contribution.}\qquad Guangyu Tang\thanks{\texttt{tang\_guangyu@126.com}.  Equal contribution.}\qquad Jiaojiao Jiang\thanks{\texttt{jiaojiao.jiang@unsw.edu.au}. Corresponding author. }\\
School of Computer Science and Engineering \\
University of New South Wales \\
Sydney, NSW 2052, Australia
}

\else

\fi

\ifdefined\isarxiv
\begin{titlepage}
  \maketitle
  \begin{abstract}
Many real-world interactions are group-based rather than pairwise such as papers with multiple co-authors and users jointly engaging with items. Hypergraph neural networks have shown great promise at modeling higher-order relations, but their reliance on a fixed number of explicit message-passing layers limits long-range dependency capture and can destabilize training as depth grows. 
In this work, we introduce Implicit Hypergraph Neural Networks (IHGNN), which bring the implicit equilibrium formulation to hypergraphs: instead of stacking layers, IHGNN computes representations as the solution to a nonlinear fixed-point equation, enabling stable and efficient global propagation across hyperedges without deep architectures.
We develop a well-posed training scheme with provable convergence, analyze the oversmoothing conditions and expressivity of the model, and derive a transductive generalization bound on hypergraphs. We further present an implicit-gradient training procedure coupled with a projection-based stabilization strategy. Extensive experiments on citation benchmarks show that IHGNN consistently outperforms strong traditional graph/hypergraph neural network baselines in both accuracy and robustness. Empirically, IHGNN is resilient to random initialization and hyperparameter variation, highlighting its strong generalization and practical value for higher-order relational learning.

  \end{abstract}
  \thispagestyle{empty}
\end{titlepage}

{\hypersetup{linkcolor={black}}
\tableofcontents
}
\newpage

\else

\begin{abstract}
Many real-world interactions are group-based rather than pairwise such as papers with multiple co-authors and users jointly engaging with items. Hypergraph neural networks have shown great promise at modeling higher-order relations, but their reliance on a fixed number of explicit message-passing layers limits long-range dependency capture and can destabilize training as depth grows. 
In this work, we introduce Implicit Hypergraph Neural Networks (IHGNN), which bring the implicit equilibrium formulation to hypergraphs: instead of stacking layers, IHGNN computes representations as the solution to a nonlinear fixed-point equation, enabling stable and efficient global propagation across hyperedges without deep architectures.
We develop a well-posed training scheme with provable convergence, analyze the oversmoothing conditions and expressivity of the model, and derive a transductive generalization bound on hypergraphs. We further present an implicit-gradient training procedure coupled with a projection-based stabilization strategy. Extensive experiments on citation benchmarks show that IHGNN consistently outperforms strong traditional graph/hypergraph neural network baselines in both accuracy and robustness. Empirically, IHGNN is resilient to random initialization and hyperparameter variation, highlighting its strong generalization and practical value for higher-order relational learning.
\end{abstract}

\fi

\section{Introduction}

Graph neural networks (GNNs) have emerged as a powerful paradigm for learning from graph-structured data, where nodes represent entities and edges capture their pairwise relationships~\citep{kipf2017gcn,wu2020comprehensive,wu2022graph}. Since graph structures naturally appear in a wide range of real-world scenarios, GNNs have been successfully applied across diverse domains, including social network analysis~\citep{jain2023opinion,zhang2024doubleh}, recommendation systems~\citep{fan2020graph,wu2022graph,sharma2024survey}, fake news detection~\citep{xie2023knowledge,gong2023fake,chang2024dihan}, predictive healthcare~\citep{hu2024bridging,hu2025udoncare,han2025no}, and detecting software vulnerabilities~\citep{cheng2019static,cheng2021deepwukong,chu2024graphneural}. These applications highlight the generality and adaptability of GNNs in modeling pairwise relations across different modalities and contexts.

However, many real-world scenarios involve complex, higher-order interactions that cannot be fully captured by simple pairwise connections. For example, in a coauthorship network~\citep{kumar2015co}, a single paper often involves more than two authors. Such relationships are more naturally represented using a hypergraph, in which each vertex corresponds to an author and each hyperedge connects all authors of the same paper. By explicitly modeling these multi-way correlations, hypergraphs can capture intricate interdependencies among multiple entities simultaneously, providing a richer and more faithful representation of complex relationships present in real-world data.

Hypergraph neural networks (HGNNs) naturally generalize GNNs to learn from hypergraph-structured data~\citep{feng2019hgnn,yadati2019hypergcn,jiang2019dynamic,gao2022hgnn+}. By extending the capabilities of GNNs, HGNNs can flexibly model and analyze complex, higher-order relationships that arise in many domains. This expressive power has fueled growing research interest and enabled successful applications in diverse areas, including computer vision~\citep{sasikaladevi2022hypergraph,han2023vision,ma2024multi,lei2025softhgnn} and natural language processing~\citep{ding2020more,bazaga2024hyperbert,feng2025beyond}, which underscore the versatility of HGNNs in domains where higher-order structure is crucial for accurate modeling.

Despite their expressive power, conventional HGNNs still rely on explicit message passing across stacked layers, which is inherently limited in capturing long-range dependencies. As depth increases, training becomes prone to vanishing or exploding gradients~\citep{he2016resnet,han2021resgcn}, and models often suffer from computational inefficiency and instability. Moreover, oversmoothing~\citep{li2018deeper,oono2020graph,chen2022preventing}, where node representations become indistinguishable as layers stack, can severely degrade performance, particularly in tasks requiring fine-grained discrimination. These challenges highlight the need for more effective architectures that can capture global context without sacrificing stability or efficiency. 

To address the above challenges, we draw inspiration from the success of implicit models~\citep{chen2018neural,bai2019deep,gu2020ignn,fu2023implicit}, which compute feature representations by solving nonlinear fixed-point equations rather than propagating information through stacked message-passing layers. \citet{gu2020ignn} applied this paradigm to GNNs, enabling the capture of long-range dependencies in graphs. However, its extension to hypergraph neural networks remains unexplored. To fill this gap, we propose the 
\textbf{\underline{I}mplicit \underline{H}yper\underline{g}raph \underline{N}eural \underline{N}etwork} (\textbf{IHGNN}), which enjoys both the expressive power of hypergraph modeling with the stability and depth efficiency and stability of implicit architectures. IHGNN performs global reasoning in a single step by directly solving a nonlinear fixed-point equation, effectively capturing higher-order, long-range dependencies while avoiding the instability and inefficiency of deep stacked models. We provide a comprehensive theoretical analysis and extensive empirical evaluation to demonstrate its effectiveness.

Below, we summarize the main contributions of this work:
\begin{itemize}
\item \textbf{Implicit Hypergraph Learning Framework:} We propose the first hypergraph neural architecture that integrates implicit equilibrium formulations, enabling expressive representation learning without layer-wise message-passing iterations.
\item \textbf{Comprehensive Theoretical Analysis:} We establish a well-posed training scheme with provable convergence guarantees, theoretically show that IHGNN mitigates oversmoothing, and derive a generalization bound for transductive learning on hypergraphs.
\item \textbf{Empirical Evaluation:} On Cora, Pubmed, and Citeseer citation benchmarks, IHGNN achieves state-of-the-art performance, exhibiting robust accuracy, parameter stability, and training resilience under diverse conditions.
\end{itemize}

\paragraph{Roadmap}
In \cref{sec:related}, we survey related work on GNNs/HGNNs, implicit graph models, and graph neural ODEs.
In \cref{sec:preli}, we present the preliminaries and notation used throughout.
In \cref{sec:methodology}, we detail the IHGNN architecture and our theoretical analysis.
In \cref{sec:experiemnt}, we report experimental results and ablations.
In \cref{sec:conclu}, we conclude and discuss future directions.
\section{Related Work}\label{sec:related}

\subsection{Graph and Hypergraph Neural Networks}

Graph neural networks (GNNs) have become a foundational approach for learning over graph‑structured data. Classical GNN models introduced neighborhood aggregation and attention mechanisms that achieved strong performance on node‑level tasks~\citep{kipf2017gcn,hamilton2017graphsage,velickovic2018gat,xu2019gin,li2019deepergcn}. Despite their successes, standard message-passing graph neural networks are fundamentally restricted to modeling pairwise relationships between nodes. As network depth increases, they often suffer from over-smoothing, i.e., node representations become indistinguishable, and limited receptive fields, which hinder their ability to capture long-range dependencies~\citep{oono2020graph,bodnar2023neural,shi2023exposition,shao2023unifying,rusch2023survey}.

To capture higher‑order structure information beyond pairwise edges, \citet{feng2019hypergraph} developed HGNN which propagates signals along node–hyperedge–node paths using an incidence-based Laplacian. \citet{bai2021hypergraph_conv} studied hypergraph convolution and hypergraph attention networks. \citet{dong2020hnhn} introduced HNHN that leverages explicit hyperedge neurons with nonlinearities and degree/cardinality-aware normalization. \citep{ding2020more} proposed HyperGAT, a hypergraph attention networks for inductive text classification. HyperFormer~\citep{ding2023hyperformer} is a hyper-relational knowledge graph completer with mixture-of-experts layers for better accuracy at lower cost. In addition, \citet{ijcai2019dynamic} extend the idea to dynamic hypergraphs by alternating topology construction with hypergraph convolution so the structure adapts during learning, and \citet{cai2022hypergraph} introduced the hypergraph structure learning for hypergraph neural networks. \citet{huang2021unignn} established a unified framework for graph neural networks and hypergraph neural networks.

These models have been adapted to a variety of domains including recommendation~\citep{li2022hyperrec,khan2025heterogeneous}, multi‑modal learning~\citep{kim2020hypergraph,huang2024dynamic}, and social network analysis~\citep{taheri2021socialhyper}. More recent works on HGNNs include~\cite{li2025deep,yang2025recent,xie2025k,feng2025kernelized,ji2025mode}. However, most HGNNs still rely on explicit iterative message passing, which becomes computationally expensive and can be unstable as depth grows.

\subsection{Implicit Graph Models and Graph Neural ODEs}

Implicit models originate from the idea of replacing explicit layer stacking with the solution of an equilibrium or continuous-depth system. Deep equilibrium models recast an infinitely deep network as a root-finding problem whose fixed point defines the representation; they train end-to-end via implicit differentiation, yielding constant memory (up to solver tolerance) and facilitating long-range dependency modeling~\citep{bai2019deep}. In parallel, Neural ODEs view depth as continuous time and use ODE solvers for forward inference together with adjoint-based or implicit differentiation for training, mitigating optimization issues associated with very deep stacks~\citep{chen2018neural,dupont2019augmented}. Both lines share a common premise: use a solver instead of layers to achieve global information propagation, better memory efficiency, and improved stability.

Building on these foundations, implicit graph models transfer the equilibrium/continuous-depth perspective to graph-structured data. \citet{gu2020ignn} introduced IGNN by formulating graph inference as solving a nonlinear fixed-point equation so that information couples across the entire graph without deep propagation. Later, \cite{baker2023implicit} leverage the monotone operator theory and enhance the performance of IGNN in learning long range dependencies. \citet{liu2022mgnni} developed multiscale graph neural networks with implicit layers. \citet{fu2023implicit} introduced a framework for designing implicit graph diffusion layers using parameterized graph Laplacians. \citet{liu2024scalable} showed how to efficiently train implicit GNNs to provide effective predictions on large graphs. \citet{lin2024ignn} introduced a learnable neural solve for IGNNs. \cite{han2024continuous} identifies the core building blocks for adapting continuous dynamics to GNNs and proposes a general framework for designing graph neural dynamics. The recent work of \citet{yang2025implicit} systematically compared implicit and unfolded GNNs from both empirical and theoretical perspectives.

From the continuous-depth side, Graph Neural ODEs model the evolution of node states along a learned vector field on the graph, often yielding smoother gradients and robust training on large or sparse graphs~\citep{poli2019graph,xhonneux2020continuous,jin2022multivariate,liu2025graph}.

\section{Preliminaries}\label{sec:preli}

In this section, we first define the notation used throughout the paper and then provide background on hypergraphs and hypergraph neural networks.

\subsection{Notation}

For any $n \in \mathbb N$, we use $[n]$ to denote the set $\{1,2,\ldots,n\}$.  We denote vectors and matrices by lower- and upper-case boldface letters, respectively. 
For a vector $\bx \in \R^d$, we use $\|\bx\|_1, \|\bx\|_2$, and $\|\bx\|_\infty$ to denote the $\ell_1$-, $\ell_2$-, and $\ell_\infty$-norm of $\bx$. We use $\mathbf{0}_n$ and $\mathbf{1}_n$ to denote an $n$-dimensional all-zero and all-one vectors, respectively. For two vectors $\bx, \by \in \R^n$, we use $\langle \bx, \by \rangle$ or $\bx^\top \by$ to denote their standard inner product. Given a vector $\bx \in \R^d$, let $\Diag(\bx)$ denote the diagonal matrix with $\Diag(\bx)_{i,i} = \bx_i$ for $i \in [d]$ and zeros elsewhere.

For a matrix $\bA \in \R^{n \times d}$, we use $\bA_{i,j}$ to denote its $(i,j)$-th entry and use $\bA^\top$ to denote its transpose. The largest eigenvalue of $\bA$ is denoted as $\lambda_{\max}(\bA)$. For two matrices $\bA \in \R^{n_1 \times d_1}$ and $\bB \in \R^{n_2 \times d_2}$, we denote their Kronecker product as $\bA \otimes \bB$ where $(\bA \otimes \bB)_{(i_1-1)n_2+i_2, (j_1-1)d_2+j_2} := \bA_{i_1, j_1} \bB_{i_2, j_2}$ for $i_1 \in [n_1], j_1 \in [d_1], i_2 \in [n_2], j_2 \in [d_2]$. 
For two matrices $\bA, \bB \in \R^{m \times n}$ of the same size, we denote their Hardmard product as $\bA \odot \bB$ where $(\bA \odot \bB)_{i,j} = \bA_{i,j}\bB_{i,j}$. 
For a matrix $\bA \in \R^{n\times d}$, its vectorization is defined as $\vect(\bA) \in \R^{nd}$ where $\vect(\bA)_{(i-1)d+j} := \bA_{i,j}$ for $i \in [n], j \in [d]$. For a matrix $\bA \in \R^{n\times d}$, we use $\|\bA\|_F$ to denote its Frobienus norm, $\|\bA\|$ to denote its spectral norm, and $\| \bA \|_\infty := \max_{i \in [n]} \sum_{j \in [d]} |\bA_{i,j}|$ to denote its maximum-row-sum norm. For two matrices $\bA, \bB \in \R^{n \times d}$, we define the matrix inner product $\langle \bA, \bB\rangle := \tr[\bA\bB^\top]$.

\subsection{Hypergraph}
A (weighted) hypergraph is defined as $G = (V,E, \bw)$, which contains a set of nodes $V = \{v_1, v_2, \ldots, v_{|V|}\}$, a set of hyperedges $E = \{e_1, \ldots, e_{|E|}\}$, and a hyperedge-weight vector $\bw = [w_1, \ldots, w_{|E|}]^\top \in \R^{|E|}$. Each hyperedge $e_j$ is a nonempty subset of nodes and is assigned with a weight $w_j$. We denote $n := |V|$ as the number of nodes and $m := |E|$ as the number of hyperedges. We can represent the set $E$ as an incidence matrix $\bH \in \{0,1\}^{|V| \times |E|}$, where for every $i \in [n], j \in [m]$, 
\begin{align*}
    \bH_{i,j} := 
    \begin{cases}
        1, & \mathrm{if~} v_i \in e_j, \\
        0, & \mathrm{otherwise}.
    \end{cases}
\end{align*}

The hyperedge weight matrix $\bE := \Diag(\bw) \in \R^{m \times m}$ is defined as a diagonal matrix with $\bE_{j,j} := w_j$ for each $j \in [m]$. The node degree matrix $\bD := \Diag(\bH\bw) \in \R^{n \times n}$ is defined as a diagonal matrix with $\bD_{i,i} := \sum_{j=1}^m \bH_{i,j} w_{i,j}$ for each $i \in [n]$. The hyperedge degree matrix $\bB := \Diag(\bH^\top \mathbf{1}_m) \in \R^{m \times m}$ is defined as a diagonal matrix with $\bB_{j,j} := \sum_{i=1}^n \bH_{i,j}$ for each $j \in [m]$.

\subsection{Hypergraph Neural Networks}

We assume that each node $v_i$ is equipped with input node feature $\bx_i \in \R^d$. We denote the input node feature matrix as 
\begin{align*}
    \bX := \begin{bmatrix}
        \bx_1 & \bx_2 & \ldots & \bx_{n}
    \end{bmatrix}^\top \in \R^{n \times d}
\end{align*}
where the $i$-th row of $\bX$ is node feature $\bx_i$. 

In the traditional Hypergraph Neural Network (HGNN) framework~\citep{feng2019hypergraph,bai2021hypergraph_conv}, node features are updated iteratively through explicit layer-wise propagation. Formally, the $t$-th layer of an HGNN is defined as
\begin{align*}
\bX^{(t+1)} = &~ f^{(t)}(\bX^{(t)}; \bW) \\
= &~ \phi\big(\bD^{-1/2} \bH \bE \bB^{-1} \bH^{\top} \bD^{-1/2} \bX^{(t)} \bW \big),
\end{align*}
where $\bW \in \R^{d \times d}$ is the trainable weight matrix, $\bX^{(t)}$ is the input feature matrix at the $t$-th layer with $\bX^{(1)} := \bX$, $\phi:\R \to \R$ is an entry-wise nonlinear activation function, and  $\bD^{-1/2} \bH \bE \bB^{-1} \bH^{\top} \bD^{-1/2}$ is the normalized hypergraph Laplacian matrix that governs feature propagation on the hypergraph. 
In other words, the normalized hypergraph Laplacian matrix serves as the propagation operator in HGNN, 
allowing information to flow across multiple nodes connected by common hyperedges and thus enabling the learning of high-order relationships. Note that when a hypergraph degenerates to a graph, it is exactly the normalized Laplacian matrix for the graph.

\section{Implicity Hypergraph Neural Networks}\label{sec:methodology}

This section presents the construction and theoretical foundations of the proposed IHGNN. We begin by reviewing the structure and spectral convolution principles of traditional HGNNs. We then introduce the integration of nonlinear equilibrium equations from IGNNs into the hypergraph setting, forming the basis of IHGNN. This integration enables more effective modeling of complex high-order dependencies while improving computational stability and convergence. Finally, we detail the well-posedness analysis and optimization strategy employed during training.

\subsection{The Architecture of IHGNN}

Traditional HGNNs rely on explicit layer-wise propagation over a fixed number of iterations to perform feature aggregation. While effective, this approach often struggles to capture long-range dependencies and can suffer from instability in deep architectures. To address these limitations, we propose the Implicit Hypergraph Neural Network (IHGNN), which incorporates the nonlinear equilibrium formulation from implicit graph neural networks~\citep{gu2020ignn} into the hypergraph setting. This integration preserves the high-order relational modeling capabilities of HGNNs while enabling more stable and efficient representation learning through implicit computation.

\paragraph{IHGNN Model}

In IHGNN, the node representations are computed by solving a nonlinear fixed-point equation, rather than through iterative layer-wise updates. This equilibrium-based formulation allows for the modeling of global dependencies without increasing model depth, thereby improving stability and scalability. We formally define the IHGNN as follows.

\begin{definition}[IHGNN] 
The architecture of IHGNN  is defined as the following mapping
\begin{align*}
    \hat{\bY} = f(\bX; \bW, \bTheta_1, \bTheta_2, \bb)
\end{align*}
where $\bX \in \R^{n \times d}$ is the input node feature matrix, $\bW \in \R^{d_h \times d_h}, \bTheta_1 \in \R^{d \times d_h} , \bb \in \R^{d_h}$, $\bTheta_2 \in \R^{d_h \times d'}$ are trainable weights, and $f$ can be described with the following equations:
\begin{align}
  \wt{\bX} = &~ \bX \bTheta_1  + \mathbf{1}_n \bb^\top, \label{eq:ihgnn_aff} \\
    \bZ  = &~ \phi\left(\bD^{-1/2} \bH \bE \bB^{-1} \bH^{\top} \bD^{-1/2} \bZ \bW+ \wt{\bX} \right), \label{eq:ihgnn_equ} \\
    \hat{\bY} = &~ \bZ \bTheta_2, \label{eq:ihgnn_output}
\end{align}
where $\phi:\R\to\R$ is a nonlinear activation function.
\end{definition}

Now we introduce the components of the IHGNN. \cref{eq:ihgnn_aff} defines an affine transformation that serves as a feature–preprocessing unit, injecting a learned skip term into the equilibrium layer, and \cref{eq:ihgnn_output} produces the final prediction via a linear readout on the equilibrium state. The implicit layer \cref{eq:ihgnn_equ} can be viewed as solving the fixed-point equation
\begin{align*}
    \bZ  = \cT(\bZ) := \phi\left(\bD^{-1/2} \bH \bE \bB^{-1} \bH^{\top} \bD^{-1/2} \bZ \bW+ \wt{\bX} \right).
\end{align*}
Intuitively, $\bD^{-1/2} \bH \bE \bB^{-1} \bH^{\top} \bD^{-1/2}$ is a hypergraph propogate operator which aggregates along node–hyperedge–node paths, so solving $\bZ = \cT(\bZ)$ performs global, higher-order reasoning in a single equilibrium step rather than by stacking many layers. We will show that under some mild conditions, $\bZ  = \cT(\bZ)$ has a unique solution $\bZ^*$, and the fixed-point iteration $\bZ^{(t+1)}  = \cT(\bZ^{(t)})$ converges to $\bZ^*$ as $t \to \infty$. Hence $f$ is a well-defined mapping in such cases.

The implicit formulation mitigates common issues in deep neural networks such as vanishing or exploding gradients. By stabilizing node states through convergence to a fixed point, IHGNN supports consistent and robust learning, particularly in large and complex hypergraph structures where traditional iterative models may become unstable.


\subsection{Well-Posedness and Convergence Analysis}

To ensure that the IHGNN model produces valid and stable node representations, it is essential to guarantee the existence and uniqueness of a solution to the implicit equilibrium equation for any given input. In certain cases, such a solution may not exist or may not be unique. Therefore, establishing the well-posedness of the model is crucial for both theoretical soundness and practical convergence.

To simplify the notation, we denote the normalized hypergraph Laplacian matrix as
\begin{align*}
    \bM := \bD^{-1/2} \bH \bE \bB^{-1} \bH^{\top} \bD^{-1/2}.
\end{align*}
Then the fixed-point equilibrium equation ,i.e., \cref{eq:ihgnn_equ}, in IHGNN becomes
\begin{align}\label{eq:fix_iter_sim}
\bZ = \phi\left(\bM\bZ\bW + \wt{\bX}\right).
\end{align}
We say that \cref{eq:fix_iter_sim} is well-posed if for any $\wt{\bX} \in \R^{d \times d}$, it has a unique solution $\bZ^*$.


To derive a sufficient condition for the well-posedness of the fixed-point equilibrium equation. We assume that the IHGNN is constructed on an admissible hypergraph, which is defined as follows.

\begin{definition}[Admissible hypergraph]
We say that a hypergraph is admissible if each hyperedge is associated with a non‑negative weight, and each node has a positive degree.
\end{definition}

Nonnegative hyperedge weights and strictly positive node degrees ensure the normalized hypergraph Laplacian $\bM = \bD^{-1/2} \bH \bE \bB^{-1} \bH^{\top} \bD^{-1/2}$ is well defined, since $\bE \succeq 0$ and $\bD^{-1/2}$ and $\bB^{-1}$ exist. Under these conditions $\bM$ is positive semidefinite. Combined with the Lipschitz continuity of activation function $\phi$ and a spectral norm bound on $\bW$, this yields a simple well-posedness condition. The proof is defered to \cref{app:well-posed}.

\begin{theorem}[Sufficient condition for well-posedness]\label{thm:pf_hypergraph}
Let $\bM \in \R^{n \times n}$ be the normalized hypergraph Laplcaian matrix of an admissible hypergraph. Assume that $\phi:\R\to\R$ is a nonexpansive activation function, i.e., $\phi$ is $1$-Lipschit. If the weight matrix $\bW \in \R^{d_h \times d_h}$ of an IHGNN satisfies $\lambda_{\max}(|\bW|) < 1,$ then for any $\wt{\bX} \in \R^{d\times d}$, the fixed-point equilibrium equation
\begin{align*}
    \bZ = \phi\left(\bM\bZ\bW + \wt{\bX}\right)
\end{align*}
has a unique solution $\bZ^* \in \R^{n \times d}$, and the fixed point iteration
\begin{align*}\label{eq:fix_iter_sim}
\bZ^{(t+1)} = \phi\left(\bM\bZ^{(t)}\bW + \wt{\bX}\right)
\end{align*}
converges to $\bZ^*$ as $t \to \infty$. Futhermore, if we assume that $\|\bZ^*\|_F \leq C_0$ for some $C_0 > 0$, $\lambda_{\max}(|\bW|) = \kappa$ for some $\kappa \in [0,1)$, and $\bZ^{(1)} = \mathbf{0}_{n \times d}$, then for any integer $t \geq 1$,
\begin{align*}
    \|\bZ^{(t)} - \bZ^* \|_F \leq  \kappa^{t-1} C_0.
\end{align*}
\end{theorem}

\begin{remark}[Initialization condition]
    In practice, the initialization $\bZ^{(1)}$ may be chosen arbitrarily, provided it starts within some Frobenius-norm neighborhood of the equilibrium solution $\bZ^*$, say $\|\bZ^{(1)}-\bZ^*\|_F \le C_1$ for some $C_1 \geq 0$. Under this assumption, Theorem~\ref{thm:pf_hypergraph} still guarantees geometric convergence, and the bound become $\|\bZ^{(t)} - \bZ^* \|_F \leq  \kappa^{t-1} C_1$.
\end{remark}

\begin{remark}[Comparison with prior work]
In \cite{gu2020ignn}, they studies the fixed-point layer $\bZ=\phi(\bA\bZ\bW+\mathrm{bias})$ on \emph{graphs} with adjacency matrix $\bA$, and proves well-posedness under the spectral condition of
$\bA\otimes \bW$, i.e., a joint constraint that depends on both the graph structure and the weight matrix. 
In contrast, Theorem~\ref{thm:pf_hypergraph} is stated for \emph{hypergraps} and uses the normalized hypergraph operator $\bM$.
This normalization collapses the joint constraint into the graph-agnostic requirement $\lambda_{\max}(|\bW|)<1$, yielding existence, uniqueness, and geometric convergence with rate $\kappa=\lambda_{\max}(|\bW|)$. Moreover, employing a normalized Laplacian rather than the raw adjacency is standard and more realistic in practice: it controls the spectrum, mitigates degree heterogeneity, and avoids graph-dependent spectral blow-up that would otherwise tighten the contraction bound.
\end{remark}

\subsection{Oversmoothing Analysis}

Oversmoothing, i.e., the tendency of node embeddings to collapse toward an indistinguishable constant vector as depth increases—remains a central obstacle for deep (hyper)graph networks. Since IHGNN is an implicit architecture, depth is replaced by a fixed-point‐solving procedure. Consequently, the classical layer-wise view of oversmoothing no longer applies directly. 




Our first result is a sufficient condition for IHGNN to provably avoid the trivial constant solution.
\begin{theorem}[Sufficient condition for nonidentical node features]\label{thm:nonid_node_feature}
    Let $\bM \in \R^{n \times n}$ be the normalized hypergraph Laplcaian matrix of an admissible hypergraph.  Let $\phi:\R\to\R$ be a strictly increasing nonexpansive activation function. Suppose that the weight matrix $\bW \in \R^{d \times d}$ of an IHGNN satisfies $\lambda_{\max}(|\bW|) < 1,$ then for any $\wt{\bX} \in \R^{d\times d}$ satisfying $\bx_i \neq \bx_j$ for some $i, j \in [n]$, there does not exists $\bz_0 \in \R^d$, such that $\bZ^* = \mathbf{1}_n\bz_0^\top$.
\end{theorem}


Our next result complements the previous one by showing that, even under the identity activation, IHGNN remains as expressive as any $K$-th-order polynomial hypergraph filter. This extends the spectral analysis of explicit graph models~\cite{chen2022preventing} to the implicit setting.

\begin{theorem}[Expressivity of IHGNN]\label{thm:expressivity}
    Let $\bM \in \R^{n \times n}$ be the normalized hypergraph Laplcaian matrix of an admissible hypergraph. Let $K \in \N$. For every $K$-order polynomial filter function $p(\bX) := (\sum_{k=0}^K \theta_k \bM^k)\bX$ with arbitrary coefficients $\{\theta_k\}_{k=0}^K$ and input feature matrix $\bx \in \R^{n \times d}$, there exists an IHGNN with identity activation can express it.
\end{theorem}

The proofs of above theorems are in \cref{app:over_smooth}.

\subsection{Transductive Generalization Analysis}

In this section, we conduct a theoretical analysis of transductive learning on hypergraphs, which extends this setting in graphs~\citep{fu2023implicit}. Let $\cX := \R^d$ be the input feature space and $\cY := \R^{d'}$ be the output label space. 
In the transductive setting, we observe the entire hypergraph $G$ and node features $\{\bx_i\}_{i=1}^{n}$, but labels only for a subset of nodes. 
Let $S\subseteq[n]$ denote the labeled indices with $|S|=s$ and $U=[n]\setminus S$ the unlabeled indices with $|U|=u$ (so $s+u=n$). 
During training, the learner has access to $\{\bx_i\}_{i=1}^{n}$ and $\{\by_i\}_{i\in S}$, and the goal is to predict $\{\by_i\}_{i\in U}$, i.e., the labels for all nodes with inidices in $U$. 
Without loss of generality, we index nodes so that $S=\{1,\dots,s\}$ and $U=\{s+1,\dots,n\}$.
For any $f \in \cH$, we define the training and testing as follows:
\begin{align*}
    \mathrm{Training~loss:}~~\wh{\cL}_s(f) := &~ \frac{1}{s}\sum_{i=1}^s\ell(f(\bx_i), \by_i), \\
    \mathrm{Testing~loss:}~~\cL_u(f) := &~ \frac{1}{u}\sum_{i=s+1}^{s+u}\ell(f(\bx_i), \by_i),
\end{align*}
where $\ell: \cH \times \cX \times \cY \to [0, \infty)$ is a loss function.

\begin{assumption}\label{as:ass_generalization}
     We assume the following conditions hold.
     \begin{itemize}
         \item \textbf{Bounded input features}: The input node feature matrix $\bX \in \R^{n \times d}$ satisfies $\|\bx_i\|_2 \leq C_X$ for each $i \in [n]$ for some $C_X \geq 0$.
         \item \textbf{Bounded trainable parameters}: The trainable parameters $\bTheta_1, \bTheta_2, \bb, \bW$ satisfies $\|\bTheta_1\|_F \leq \rho_1, \|\bTheta_2\|_F \leq \rho_2, \|\bb\|_2 \leq C_b$ for some $\rho_1, \rho_2, C_b \geq 0$, and $\|\bW\| \leq \kappa$ for some $\kappa \in [0,1)$. For simplicity, we assume their dimensions satisfies $d = d_h = d'$.
         \item \textbf{Lipschitz loss}: The loss function $\ell$ is $C_\ell$-Lipschitz for some $C_\ell \geq 0$.
         \item \textbf{Nonexpansive activation}: The activation function $\phi$ is nonexpansive, i.e., $1$-Lipschitz.
     \end{itemize}
\end{assumption}

We briefly argue that these assumptions are standard and easy to meet in real-world. We first dicuss the first two assumptions. Feature vectors are routinely normalized in practice, e.g., $\ell_2$-normalization, and many benchmark node features are already bounded. During the training, weight decay/regularization directly impose norm constraints on $\bTheta_1,\bTheta_2,\bb$. The spectral bound $\|\bW\|\le\kappa<1$ directly enforces the contraction needed for a unique equilibrium when combined with the normalized operator so the fixed point exists and is reached geometrically. 

For the last two assumptions, Lipschitz loss and nonexpansive activations are very common and easy to meet this on bounded domains, e.g., squared/hinge losses, cross-entropy is Lipschitz in the probability simplex or when logits are bounded. Many activations are $1$-Lipschitz (ReLU, leaky-ReLU with slope $\le1$, $\tanh$, sigmoid/softplus).

Next, we state our main result on transductive generalization bound of IHGNN. The proofs of this section are deferred to \cref{app:gen_bound}.

\begin{theorem}[Transductive generalization bound of IHGNN]\label{thm:gen_bound_IHGNN}
    We assume that the hypergraph is admissible and \cref{as:ass_generalization} are satisfied. Let $\cH$ be the hypothesis class of IHGNN models defined on the given hypergraph. Let $c_0 := \sqrt{\frac{32\log(4e)}{3}} < 5.05$.
     Let $P := \frac{1}{s} + \frac{1}{u}$, and $Q := \frac{s+u}{(s+u-1/2)(1-1/(2\max\{s,u\}))}.$ Then, for any $\delta > 0$, with
    probability at least $1-\delta$ over the choice of the training set $\{\bx_i\}_{i=1}^{s+u} \cup \{y_i\}_{i=1}^{s}$, for all $f \in \cH$, we have
    \begin{align*}
    \cL_u(f) \leq &~ \wh{\cL}_s(f) + \frac{\sqrt{2}\rho_2 C_\ell(\rho_1 C_x + \sqrt{d}C_b)}{(1-\kappa)\sqrt{s+u}} + c_0 P \sqrt{\min\{s,u\}}+ \sqrt{\frac{PQ}{2}\log\frac{1}{\delta}}.
    \end{align*}
\end{theorem}

In the asymptotic regime, we further simplify the generalization bound.

\begin{corollary}[Asymptotic transductive generalization bound of IHGNN]\label{cor:asy_gen_bound_IHGNN}
    Under the same conditions in~\cref{thm:gen_bound_IHGNN}.
    For sufficiently large training-set size $s$ and testing-set size $u$, for any $\delta > 0$, with
    probability at least $1-\delta$ over the choice of the training set, for all $f \in \cH$, we have
    \begin{align*}
    \cL_u(f) \leq &~ \wh{\cL}_s(f) + O\left(\frac{d}{s+u}\right)^{\frac{1}{2}} + O \left(\frac{\log(1/\delta)}{\min\{s,u\}}\right)^{\frac{1}{2}}.
    \end{align*}
\end{corollary}

\begin{remark}[Discussions on the generalization bound]
Note that the second term $O\big(\frac{d}{s+u}\big)^{1/2}$ decays as either the training-set size $s$ or the testing-set size $u$ increases. However, the last term $O\big(\frac{\log(1/\delta)}{\min\{s,u\}}\big)^{1/2}$ converges slowly whenever $s \ll u$ or $u \ll s$. The regime $s \ll u$ corresponds to an under-fitted model, whereas when $u \ll s$, the sample mean computed from the $u$ test nodes, drawn out of the $s+u$ available nodes, has high variance.
\end{remark}


\subsection{Training of IHGNN}


\paragraph{Numerically Tractable Well-Posedness Condition}

Directly enforcing the constraint $\lambda_{\max}(|\bW|) < 1$ is computationally difficult due to its non-convexity with respect to $\bW$. To address this, we introduce a more tractable surrogate constraint. By assuming the activation is positively homogeneous, meaning \( \sigma(\alpha x) = \alpha \sigma(x) \) for all \( \alpha \geq 0 \). We can derive the following condition and its proof is in~\cref{app:training}.

\begin{theorem}[Scaled Well-Posedness of IHGNN]\label{thm:scale_wellpose}
Suppose that the activation function $\phi: \R \to \R$ is positively homogeneous and nonexpansive. If an IHGNN model with weights $\bW, \bTheta_1, \bTheta_2, \bb$ satisfies $ \lambda_{\max}(|\bW|) < 1 $, then there exists an equivalent IHGNN model with weights $\wt{\bW}, \wt{\bTheta}_1, \wt{\bTheta_2}, \wt{\bb}$ such that $ \|\wt{\bW}\|_{\infty} < 1$, 
and both models produce identical outputs for all same inputs.
\end{theorem}
    

Although this constraint is stricter, it remains valid for many activation functions (e.g., ReLU, identity) that are positively homogeneous.

\paragraph{Tractable Condition for Training}

To maintain this condition during training, we apply a projection step after each gradient update:
\begin{equation}
\bW^+ = \Pi_C(W) := \argmin_{\|\bW'\|_{\infty} \leq \kappa} \|\bW' - \bW\|_F^2,
\end{equation}
where $\kappa \in [0, 1)$ is a relaxation parameter. This operation ensures that the updated weights lie within the feasible region defined by the well-posedness constraint, thereby promoting convergence and training stability throughout the learning process.

The training objective of IHGNN can be thought about as minimizing a supervised loss function $\cL (\wt{\bY}, \bY)$ that measures the discrepancy between the model predictions $\wh{\bY} $ and the ground truth labels $\bY$, while ensuring that the model remains within the well-posed region. Specifically, we impose the constraint \( \|\bW\|_{\infty} \leq \kappa \), where \( \kappa \in [0, 1) \) is a tunable relaxation parameter. Let $\Omega := (\bW, \bTheta_1, \bTheta_2, \bb)$ be the collection of trainable parameters. Then the optimization problem is formulated as:
\begin{align*}
    \min_{\Omega} &~  \cL (\wh{\bY}, \bY) \\
    \text{s.t.} &~ \bZ = \phi\left(\bM\bZ\bW + \bX\bTheta_1 + \mathbf{1}_n\bb^\top\right), \\ 
    &~ \wh{\bY} = \bZ \bTheta_2, \\
    &~ \|\bW\|_{\infty} \leq \kappa.
\end{align*}

To solve this constrained optimization problem, we adopt a projected gradient descent approach. After each gradient update, the weight matrix \( W \) is projected back onto the feasible set defined by the well-posedness constraint. This procedure ensures convergence and model stability throughout training.

\paragraph{Implicit Gradient Derivation}
The equilibrium formulation introduces implicit dependencies between parameters and outputs, requiring modified gradient computation. Using the chain rule, we compute gradients with respect to parameters $\Omega$ and the hidden state $\bZ$, followed by gradients for parameters $q \in \{\bW, \bTheta_1, \bb\}$. The gradient of the loss with respect to \( q \) is expressed as:
\begin{align*}
\nabla_{q} \cL = \left\langle \frac{\partial (\bM\bZ\bW + \bX\bTheta_1 + \mathbf{1}_n\bb^\top)}{\partial q}, \nabla_{\bZ}L \right\rangle,
\end{align*}
where $\bX$ is treated as fixed during this step. The quantity $\nabla_{\bZ} \cL$ satisfies the following recursive equation:
\begin{align*}
\nabla_{\bZ}\cL = \phi'(\bZ) \odot \left(\bM^{\top} \nabla_{\bZ}L \bW^{\top} + \nabla_{\bTheta_1}\cL+ \nabla_{\bb}\cL\right).
\end{align*}
Note that our previous analysis can be adapted to show that this equation can be efficiently approximately solved via fixed-point iteration, assuming that $\bW$ satisfies the well-posedness condition.



\section{Experiment}\label{sec:experiemnt}

\subsection{Experimental Setup}

We implement the proposed IHGNN model using the PyTorch framework. All experiments are conducted on a workstation equipped with an Intel i5-12600KF CPU and an NVIDIA RTX 4070 GPU, running Windows 11. The development environment includes Python 3.12 and Visual Studio Code, ensuring both implementation efficiency and experimental reproducibility.

\paragraph{Datasets}

To evaluate the generalization and classification performance of IHGNN, we utilize three widely adopted benchmark datasets from citation networks: \textbf{Cora} \citep{sen2008collective}, \textbf{Pubmed} \citep{namata2012query}, and \textbf{Citeseer} \citep{sen2008collective}. Table \ref{tab:dataset} shows the basic statistics of these datasets.

\begin{table}[!ht]
\centering
\caption{Datasets of used in experiments for node classification}
\label{tab:dataset}
\begin{tabular}{lcccc}
\toprule
\textbf{Dataset} & \textbf{Nodes} & \textbf{Hyper-edges} & \textbf{Features} & \textbf{Classes} \\
\midrule
Cora & 2,708 & 1,072 & 1,433 & 7 \\
Pubmed & 19,717 & 7,963 & 500 & 3 \\
Citeseer & 3,327 & 1,079 & 3,703 & 6 \\
\bottomrule
\end{tabular}
\end{table}


\paragraph{Hypergraph Construction}

To better capture high-order relationships among documents, we construct hypergraphs based on citation patterns. Each publication is modeled as a node, and we create hyperedges by grouping together all papers that cite a common target paper. Specifically, for every cited document, we collect all its citing nodes and connect them via a hyperedge. This method effectively encodes community-level citation interactions and enhances the representation of higher-order semantics beyond pairwise links.

\paragraph{Baselines}

We compare IHGNN against several representative models spanning classical and graph-based approaches for node classification:

\begin{itemize}
    \item \textbf{Semi-supervised Embedding (Semi-SE)} \citep{weston2012semi}: A semi-supervised learning method that leverages labeled and unlabeled data for node embedding.
    
    \item \textbf{DeepWalk} \citep{perozzi2014deepwalk}: An unsupervised method that learns latent representations via truncated random walks on the graph.

    \item \textbf{Planetoid} \citep{yang2016planetoid}: A semi-supervised learning framework that integrates structural and label information through joint optimization.

    \item \textbf{GCN} \citep{kipf2017gcn}: A foundational model that applies spectral graph convolutions to aggregate neighborhood information.

    \item \textbf{GAT} \citep{velickovic2018gat}: Enhances GCN by assigning dynamic attention weights to neighboring nodes based on feature relevance.

    \item \textbf{HGNN} \citep{feng2019hgnn}: Extends GCN to hypergraphs by modeling high-order relations via hyperedges.

    \item \textbf{IGNN} \citep{gu2020ignn}: Solves node embeddings through fixed-point iterations of equilibrium equations, enabling deeper propagation without explicit stacking.
\end{itemize}


\subsection{Model Comparison Analysis}

\begin{table}[t]
\centering
\caption{Comparison of model accuracy.}
\label{table:model_comparison}
\begin{tabular}{lccc}
\toprule
\textbf{Model} & \textbf{Cora (\%)} & \textbf{Pubmed (\%)} & \textbf{Citeseer (\%)} \\
\midrule
Semi-SE & 59.0 & 70.7 & 60.1 \\
DeepWalk & 67.2 & 65.3 & 43.2 \\
Planetoid & 75.7 & 77.2 & 64.7 \\
GCN & 81.5 & 79.0 & 70.3 \\
GAT & 83.0 & 79.0 & 72.5 \\
HGNN & 81.6 & 80.2 & 69.2 \\
IGNN & 84.4 & 80.3 & 73.6 \\
\hline
\textbf{IHGNN} & \textbf{85.9} & \textbf{83.8} & \textbf{75.1} \\
\bottomrule
\end{tabular}
\end{table}

\paragraph{Performance Overview}  
IHGNN consistently achieves the highest accuracy across all datasets, outperforming all baselines. Specifically, it achieves 85.9\% on Cora, 83.8\% on Pubmed, and 75.1\% on Citeseer, demonstrating its robustness and adaptability to different network structures and feature distributions. Table~\ref{table:model_comparison} reports the classification accuracy of IHGNN against baseline models.

\paragraph{Analysis of Baselines}  
Traditional embedding-based methods such as Semi-SE and DeepWalk yield lower accuracy. DeepWalk, for example, performs poorly on Citeseer (43.2\%) due to its inability to model structural dependencies beyond local random walks. Semi-SE also falls short across all datasets, highlighting its limitations in capturing global semantics.

Graph neural network baselines (GCN and GAT) show substantial improvements. GAT benefits from attention mechanisms that assign learnable weights to neighboring nodes, achieving 83.0\% on Cora and 72.5\% on Citeseer. However, both models are constrained by their reliance on pairwise edges and fail to fully leverage higher-order relationships.

HGNN and IGNN address these shortcomings by incorporating more expressive structural representations. HGNN uses hypergraphs to capture high-order dependencies, while IGNN models long-range interactions via implicit nonlinear equations. IGNN shows strong performance, especially on Cora, but lacks the flexibility of hypergraph modeling.

\paragraph{Superiority of IHGNN}  
IHGNN integrates the strengths of both IGNN and HGNN by embedding an implicit equilibrium formulation within a hypergraph framework. This hybrid design enables IHGNN to model complex higher-order and long-range dependencies more effectively. Its superior performance on Pubmed and Cora (which known for dense interconnections and thematic clustering) demonstrates this advantage.



\subsection{Stability Analysis}

\begin{table*}[t]
\centering
\caption{Stability evaluation of IHGNN over 50 trials: mean, standard deviation, and 95\% confidence intervals of F1 score and accuracy.}
\label{table:stability}
\begin{tabular}{lccccc}
\toprule
\multirow{2}{*}{\textbf{Dataset}} & \multirow{2}{*}{\textbf{Metric}} & \textbf{Mean} & \textbf{Standard} & \multicolumn{2}{c}{\textbf{95\% Confidence Interval}} \\
& & & \textbf{Deviation} & \textbf{Lower Bound} & \textbf{Upper Bound} \\
\midrule
\multirow{2}{*}{Cora} 
& F1        & 0.8621 & 0.0066 & 0.8603 & 0.8639 \\
& Accuracy  & 0.8619 & 0.0066 & 0.8601 & 0.8637 \\
\midrule
\multirow{2}{*}{Pubmed} 
& F1        & 0.8352 & 0.0028 & 0.8345 & 0.8360 \\
& Accuracy  & 0.8358 & 0.0028 & 0.8350 & 0.8366 \\
\midrule
\multirow{2}{*}{Citeseer} 
& F1        & 0.7529 & 0.0069 & 0.7510 & 0.7548 \\
& Accuracy  & 0.7508 & 0.0068 & 0.7489 & 0.7527 \\
\bottomrule
\end{tabular}
\end{table*}

To evaluate the stability of IHGNN, we performed a two-part evaluation: (1) repeated experiments under different random initializations, and (2) parameter sensitivity testing.

\paragraph{Reproducibility Analysis}

We conducted 50 independent training runs of IHGNN on the Cora, Pubmed, and Citeseer datasets using different random seeds. Table~\ref{table:stability} summarizes the mean, standard deviation, and 95\% confidence intervals for F1 score and accuracy across these trials.

Across all datasets, IHGNN exhibited exceptionally low standard deviation indicating consistent performance across runs. The 95\% confidence intervals are also narrow, reinforcing the model’s reliability and resistance to random initialization effects. For instance, on Cora, the F1 score varied only within [0.8603, 0.8639], while the accuracy varied within a narrow band of [0.8601, 0.8637].

We attribute the stability to IHGNN’s implicit equilibrium framework and hypergraph-based design, which together mitigate the stochastic variability that often affects deep learning models. Unlike traditional GNNs that can suffer from training instability and initialization sensitivity, IHGNN’s formulation ensures convergence to a fixed-point solution that captures both local and global structure in a stable manner.

Furthermore, the hypergraph structure contributes to consistent feature propagation by modeling high-order dependencies, thus reducing the model’s susceptibility to overfitting or local noise. This is particularly valuable in complex domains such as fake news detection, where even minor performance fluctuations can lead to misclassifications.


\paragraph{Hyperparameter Robustness}

To assess the robustness of IHGNN with respect to hyperparameter variations, we systematically vary the number of hidden units (\texttt{nhid}), learning rate (\texttt{lr}), and dropout rate. 
IHGNN demonstrates strong parameter stability on the Cora dataset. Across all tested settings, the model consistently achieved an average F1 score and accuracy of 0.86, with a standard deviation of just 0.01. This low variation confirms the model’s resilience to changes in architectural and training hyperparameters, ensuring dependable performance regardless of parameter tuning.

For the Pubmed dataset, the mean F1 score and accuracy were both 0.82, with a standard deviation of 0.03. Although slightly more variable than on Cora, the performance remained highly stable across all settings. This suggests that IHGNN generalizes well even when minor adjustments are made to training configurations.

On the Citeseer dataset, IHGNN achieved mean F1 and accuracy values of 0.75 and 0.76, respectively. The standard deviations (0.04 for F1, 0.03 for accuracy) were slightly higher, indicating modest sensitivity to hyperparameter choices. This may be attributed to the increased complexity and sparsity of Citeseer's graph structure, which can introduce variability in model behavior. Nonetheless, the variation remains well within acceptable bounds.

The results demonstrate that IHGNN maintains consistently high performance across a wide range of hyperparameter settings. The minimal fluctuations in F1 and accuracy, especially on Cora and Pubmed, highlight the model’s robustness and parameter insensitivity. This property is crucial for real-world applications, where exhaustive hyperparameter tuning may be impractical. By ensuring strong and stable results across different datasets and training regimes, IHGNN proves to be a reliable and adaptable model suitable for diverse classification tasks in complex networked data.

\subsection{Convergence Analysis}

\begin{figure*}[t]
    \centering
    \begin{subfigure}[t]{0.68\textwidth}
        \centering
        \includegraphics[width=\textwidth]{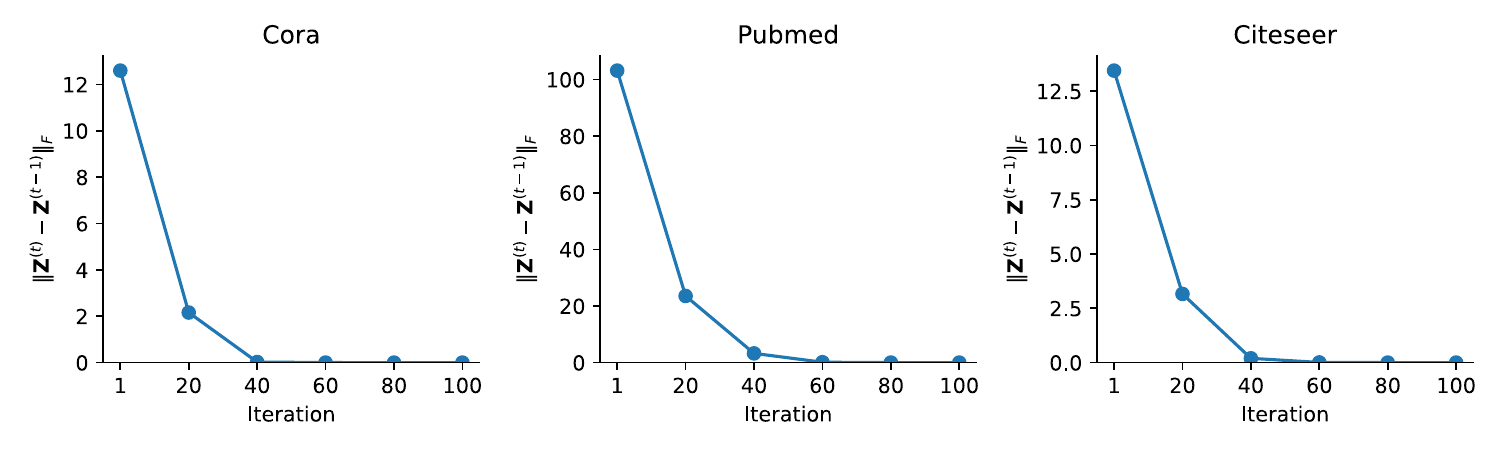}
    \end{subfigure}
    \begin{subfigure}[t]{0.31\textwidth}
        \centering
    \includegraphics[width=\textwidth]{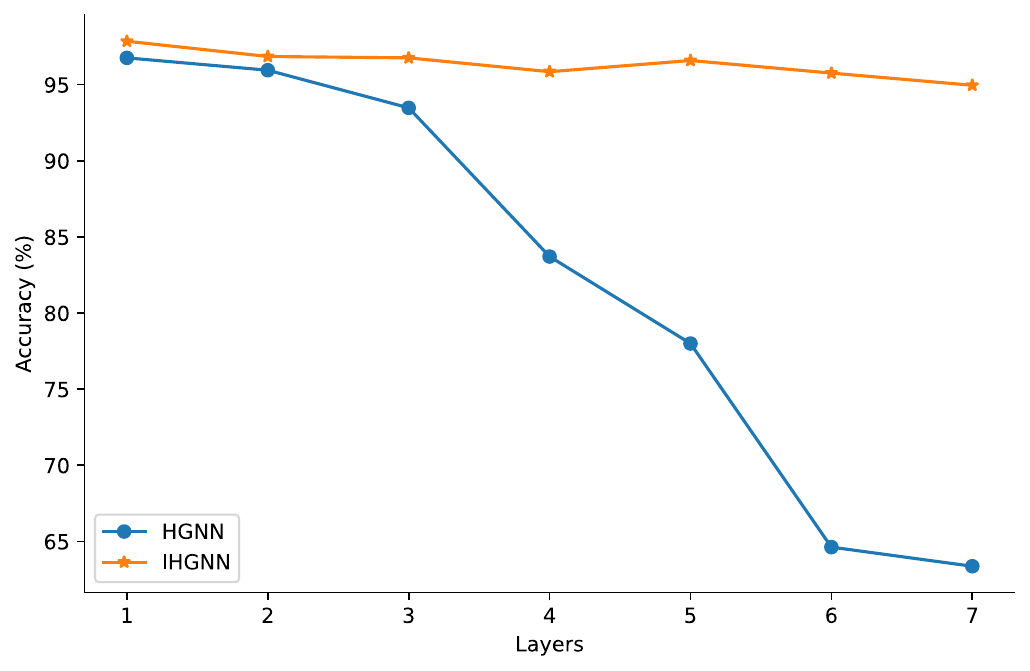}
    \end{subfigure}
    \caption{(a) Experiment results on convergence analysis of IHGNN on Cora, Pubmed, and Citeseer. (b) Comparison of oversmoothing results of HGNN and IHGNN on ModelNet40.}
    \label{fig:exp_convergence_oversmooth}
\end{figure*}

We carry out experiments on Cora, PubMed, and Citeseer datasets to investigate the convergence properties of implicit layers in IHGNN under different iteration $t$.  \cref{fig:exp_convergence_oversmooth}(a) shows that as the iteration value increases, the $\|\bZ^{(t)} - \bZ^{(t-1)} \|_F$ values in all three datasets rapidly decrease and approach 0. This verifies the convergence result of the implicit equilibrium equation.

\subsection{Oversmoothing Analysis}

To facilitate the discussion of model oversmoothing, we adopt the same experimental setup as the baseline model HGNN and conduct experiments on the ModelNet40 \citep{wu20153d} datasets. As illustrated in  \cref{fig:exp_convergence_oversmooth}(b), our model maintains robust performance with increasing layer depth, while HGNN exhibits degradation. These results clearly demonstrate IHGNN's capability to avoid over-smoothing.

\section{Conclusion and Future Work}\label{sec:conclu}


We introduced Implicit Hypergraph Neural Networks (IHGNN), a framework that unifies implicit equilibrium modeling with hypergraph architectures. By solving a single nonlinear fixed-point equation, IHGNN performs global reasoning over higher-order hyperedges without deep stacking, mitigating vanishing/exploding gradients and reducing computational overhead. Our analysis establishes well-posedness, convergence, and transductive generalization guarantees, and experiments on Cora, PubMed, and Citeseer demonstrate consistent gains in accuracy, robustness, and training stability over strong GNN/HGNN baselines. Collectively, these results show that IHGNN captures long-range, higher-order dependencies and underscore its practical value for complex relational learning, opening avenues for broader implicit modeling on non-pairwise structures.

For the future work, on the theoretical side, it is important to characterize the computational limits of hypergraph neural networks and implicit graph models, for example via circuit-complexity perspectives~\citep{grohe2024descriptive,chen2024circuit} or refined Weisfeiler–Lehman analyses~\citep{boker2023fine,wang2023mathscr,bouritsas2022improving}. On the practical side, future work could incorporate pruning and model-acceleration techniques~\citep{liu2023comprehensive,liang2025beyond} to speed up training and inference for implicit hypergraph models, and explore applications of IHGNN to fake news detection, social network analysis, and software vulnerability detection.

\ifdefined\isarxiv
\bibliographystyle{apalike}
\bibliography{ref}
\else

\fi

\newpage
\onecolumn
\appendix

\section{Preliminaries}

\subsection{Kronecker Product and Vectorization}

\begin{definition}[Kronecker product]
    Let $A \in \R^{n_1 \times d_1}$ and $B \in \R^{n_2 \times d_2}$ be two matrices. The Kronecker product of $A$ and $B$ is defined as
    \begin{align*}
        \bA \otimes \bB := \begin{bmatrix}
            \bA_{1,1}\bB & \bA_{1,2} \bB & \cdots & \bA_{1,d_1} \bB\\
            \bA_{2,1} \bB & \bA_{2,2} \bB & \cdots & \bA_{2,d_1} \bB \\
            \vdots & \vdots & \ddots & \vdots \\
            \bA_{n_1,1} \bB & \bA_{n_1,2} \bB & \cdots & \bA_{n_1,d_1} \bB
        \end{bmatrix} 
        \in \R^{n_1n_2 \times d_1d_2},
    \end{align*}
    where $(\bA \otimes \bB)_{(i_1-1)n_2 + i_2, (j_1-1)d_2 + j_2} = \bA_{i_1, j_1} \bB_{i_2, j_2}$ for $i_1 \in [n_1], j_1 \in [d_1], i_2 \in [n_2], j_2 \in [d_2]$.
\end{definition}

\begin{definition}[Vectorization]
    Let $A \in \R^{n \times d}$ be a matrix. The vectorization of $A$ is defined as
    \begin{align*}
        \vect(\bA) = \begin{bmatrix}
            \bA_{:, 1} \\
            \bA_{:, 1} \\
            \vdots \\
            \bA_{:, d}
        \end{bmatrix}
         \in \R^{n \times d}
    \end{align*}
    where $\bA_{:, j}$ is the $j$-th column of $\bA$, and $\vect(\bA)_{(i-1)d+j} = \bA_{i,j}$ for $i \in [n], j \in [d]$.
\end{definition}

\begin{lemma}[Tensor trick, Section 10.2.2 in \citet{petersen2008matrix}]\label{lem:tensor_trick}
    Given three matrices $\bA \in \R^{m \times n}, \bX \in \R^{n\times p}, \bB \in \R^{p \times q}$, we have
    \begin{align*}
        \vect(\bA \bX \bB) = (\bB^\top \otimes \bA) \vect(\bX).
    \end{align*}
\end{lemma}

\begin{lemma}[Spectrum of the Kroneck product, Theorem 4.2.12 in \citet{horn1994topics}]\label{lem:spectrum_kronecker}
    Let $\bA \in \R^{m \times n}$ and $\bB \in \R^{p \times q}$ be two matrices. Let $(\lambda, \bx)$ and $(\lambda, \bx)$ be two pairs of eigenvalue and eigenvector of $\bA$ and $\bB$, respectively. Then $\lambda\mu$ is an eigenvalue of $\bA \otimes \bB$ with corresponding eigenvector $\bx \otimes \by$. Moreover, any eigenvalue of $\bA \otimes \bB$ is a product of eigenvalues of $\bA$ and $\bB$.
\end{lemma}

\subsection{Rademacher Complexity}

\begin{definition}[Rademacher complexity, Definition 3.1 in \citet{mohri2018foundations}]
Let $\cG$ be a family of functions mapping from $\cZ$ to $\R$ and $S=\{z_i\}_{i=1}^n \subseteq \cZ$ a set of samples with elements in $\cZ$. Then, the (empirical) Rademacher complexity of $\cG$ with respect to the sample set $S$ is defined as
\begin{align*}
    \mathcal{R}_n(\cG) := \E_{\bvareps}\left[\sup_{g \in \cG} \frac{1}{n} \sum_{i=1}^n \varepsilon_i g(z_i)\right],
\end{align*}
where each $\bvareps = [\varepsilon_1, \varepsilon_2, \ldots, \varepsilon_n]^\top$, and for each $i \in [n]$, $\varepsilon_i$ is an independent Rademacher random variable, i.e., $\varepsilon_i \sim \mathsf{Uniform}(\{-1,1\})$.
\end{definition}

\begin{definition}[Coordinate-wise Rademacher complexity \citep{maurer2016vector}]
Let $\cF$ be a family of functions mapping from $\cZ$ to $\R^d$ and $S=\{z_i\}_{i=1}^n \subseteq \cZ$ a set of samples with elements in $\cZ$. Let $f(\cdot)_j$ denote the $j$-th entry of the output of the function $f(\cdot)$. Then, the (empirical) coordinate-wise Rademacher complexity of $\cF$ with respect to the sample set $S$ is defined as
\begin{align*}
    \mathcal{R}^{\mathrm{coord}}_n(\cF) := \E_{\bVareps}\left[\sup_{f \in \cF} \frac{1}{n} \sum_{i=1}^n\sum_{j=1}^n \bVareps_{i,j} f(z_i)_j\right],
\end{align*}
where each $\bVareps$ is an $n \times d$ random matrix, and for each $i \in [n], j \in [d]$, $\bVareps_{i,j}$ is an independent Rademacher random variable, i.e., $\bVareps_{i,j} \sim \mathsf{Uniform}(\{-1,1\})$.
\end{definition}

\begin{lemma}[Contraction inequality for vector-valued function class \citep{maurer2016vector}]\label{lem:contra_vect:app}
Let $\cF$ be a family of functions mapping from $\cZ$ to $\R^d$ and $S=\{z_i\}_{i=1}^n \subseteq \cZ$ a set of samples with elements in $\cZ$. Let $\rho: \R^d \to \R$ be a $K$-Lipschitz function for some $K \geq 0$. Then we have 
    \begin{align*}
        \mathcal{R}_n(\rho \circ \cF) \leq \sqrt{2} K \cdot \mathcal{R}^{\mathrm{coord}}_n(\cF).
    \end{align*}
\end{lemma}

\subsection{Transductive Rademacher Complexity}

\begin{definition}[Transductive Rademacher Complexity, Definition 1 in \citet{el2009transductive}]
Let $\cG$ be a family of functions mapping from $\cZ$ to $\R$ and $S=\{z_i\}_{i=1}^{s+u} \subseteq \cZ$ a set of samples with elements in $\cZ$.
Let $\boldsymbol{\sigma} = [\sigma_1, \sigma_2, \ldots, \sigma_n]^\top$, and for each $i \in [n]$, $\sigma_i$ is an independent random variable defined as
\begin{align*}
    \sigma_i := \begin{cases}
        1, & \mathrm{with~probability~} p, \\
        -1, & \mathrm{with~probability~} p,\\
        0, & \mathrm{with~probability~} 1-2p. \\
    \end{cases}
\end{align*}
Then, the transductive Rademacher complexity of $\cG$ with respect to the sample set $S$ is defined as
\begin{align*}
    \wt{\mathcal{R}}_{s+u}(\cG, p) := \E_{\boldsymbol{\sigma}}\left[\sup_{g \in \cG}\bigg(\frac{1}{s}+\frac{1}{u}\bigg)\sum_{i=1}^{s+u} \sigma_i g(z_i)\right].
\end{align*}
Moreover, we define $\wt{\mathcal{R}}_{s+u}(\cG) := \wt{\mathcal{R}}_{s+u}(\cG, p_0)$ with $p_0 = \frac{su}{(s+u)^2}$.
\end{definition}

\begin{lemma}[Corollary 1 in \citet{el2009transductive}]\label{lem:trans_gen_b:app}
    Let $\cH$ be a hypothesis class of functions from $\cX \to \cY$, and $\ell : \cH \times \cX \times \cY \to [0, \infty)$ be a loss function.  and $S=\{(z_i)\}_{i=1}^{s+u} \subseteq \cZ$ a set of samples with elements in $\cZ := \cX \times \cY$. Let $c_0 := \sqrt{\frac{32\log(4e)}{3}} < 5.05$. Let $P:=\frac{1}{s} + \frac{1}{u}$, and $Q:= \frac{s+u}{(s+u-1/2)(1-1/(2\max\{s,u\})}$. Then, for any $\delta \in (0,1)$, with probability at least $1-\delta$ over the random choice of $s$ training samples from $S$, for any $h \in \cH$, it holds that
    \begin{align*}
        \cL_s(f) \leq \wt{\cL}_u(f) + \wt{R}_{s+u}(\ell \circ \cH) + c_0 P \sqrt{\min\{s, u\}} + \sqrt{\frac{PQ}{2} \log \frac{1}{\delta}}.
    \end{align*}
\end{lemma}

\begin{lemma}[Lemma B.8 in \citet{fu2023implicit}]\label{def:rad_ineq:app}
    Let $\cG$ be a family of functions mapping from $\cZ$ to $\R$ and $S=\{z_i\}_{i=1}^{s+u} \subseteq \cZ$ a set of samples with elements in $\cZ$. Let $n := s+ u$. Then we have
    \begin{align*}
        \wt{\cR}_{s+u}(\cG) \leq \cR_n(\cG).
    \end{align*}
\end{lemma}

\section{Well-posedness and Convergence Analaysis}\label{app:well-posed}

\begin{definition}[Admissible hypergraph]
    We say that a hypergraph is admissible if each hyperedge is associated with a non‑negative weight, and each node has a positive degree.
\end{definition}

\begin{lemma}\label{lem:lap_eigen}
    Let $\bM$ be a hypergraph Laplacian matrix of an admissible hypergraph. Then each eigenvalue $\lambda_i$ of $\bM$ satisfies $|\lambda_i| \leq 1$. Moreover, we have $\lambda_{\max}(\bM) = 1$.
    \end{lemma}
    \begin{proof}
    We define the matrix
    \begin{align*}
    \bP := \bD^{-1} \bH \bE \bB^{-1} \bH^{\top}.
    \end{align*}
    Then we can write the normalized hypergraph Laplacian matrix as
    \begin{align*}
        \bM = \bD^{1/2}\bP \bD^{-1/2}.
    \end{align*}
    For every $i \in [n]$, by simple calculation, we have
    \begin{align*}
    \sum_{k=1}^{n} \bP_{ik}
      = &~\frac{1}{\bD_{i,i}}\sum_{j = 1}^m \bH_{i,j} w_j \cdot \frac{1}{\bB_{j,j}}
           \sum_{k = 1}^m \bH^\top_{j,k} \tag{By definitions of $\bD, \bH, \bB, \bE$} \\
      = &~ \frac{1}{\bD_{i,i}}\sum_{j = 1}^m \bH_{i,j} w_j \cdot \frac{1}{\bB_{j,j}}
           \sum_{k = 1}^m \bH_{k,j}  \tag{$\bH^\top_{j,k} = \bH_{k,j}$} \\
      = &~ \frac{1}{\bD_{i,i}}\sum_{j = 1}^m \bH_{i,j} w_j \tag{$\bB_{j,j} = \sum_{k = 1}^m \bH_{k,j}$}\\
      = &~ 1. \tag{$\bD_{i,i} = \sum_{j = 1}^m \bH_{i,j} w_j$}
    \end{align*}
    Hence every row of $\bP$ sums to $1$ and every entry of $\bP$ is nonnegative because the hypergraph is admissible. Since $\bM$ and $\bP$ have the same spectrum, it suffices to show the eigenvalues of $\bP$ satisfies the desired properties. 

    We first show that every eigenvalue $\lambda_i$ of $\bP$ satisfies $|\lambda_i| \leq 1$. Note that for any $\bx \in \R^n$,
    \begin{align}\label{eq:tmp_stoc_proof}
        \|\bP \bx\|_1 \leq \|\bP\|_\infty \|\bx\|_1 \leq \|\bx\|_1.
    \end{align}
    Let $\bv_i$ be an eigenvector associated with the eigenvalue $\lambda_i$. Note that
    \begin{align*}
        \|\bP \bv_i\|_1 = \|\lambda_i \bv_i \|_1 = |\lambda_i| \|\bv_i\|_1.
    \end{align*}
    Then by Eq.~\eqref{eq:tmp_stoc_proof}, we have $|\lambda_i|\leq 1$.

    To show that there exists some $\lambda_i = 1$, we see that
    \begin{align*}
        \bP \mathbf{1}_n = \begin{bmatrix}
            \sum_{j=1}^n \bP_{1,j} \\
            \sum_{j=1}^n \bP_{2,j} \\
            \vdots \\
            \sum_{j=1}^n \bP_{m,j}
        \end{bmatrix} = \begin{bmatrix}
            1 \\ 
            1 \\
            \vdots \\
            1
        \end{bmatrix} =
        \mathbf{1}_n.
    \end{align*}
    Thus can conclude that $\lambda_{\max} (\bP) = 1$.
\end{proof}

\begin{theorem}[Sufficient condition for well-posedness, restatement of Theorem~\ref{thm:pf_hypergraph}]\label{thm:pf_hypergraph:app}
Let $\bM \in \R^{n \times n}$ be the normalized hypergraph Laplcaian matrix of an admissible hypergraph. Assume that $\phi:\R\to\R$ is an contractive activation function, i.e., $\phi$ is $1$-Lipschitz. If the weight matrix $\bW \in \R^{d \times d}$ satisfies $\lambda_{\max}(|\bW|) := \kappa \in [0, 1)$ then for any $\wt{\bX} \in \R^{d\times d}$, the fixed-point equilibrium equation
\begin{align*}
    \bZ = \phi\left(\bM\bZ\bW + \wt{\bX}\right)
\end{align*}
has a unique solution $\bZ^* \in \R^{n \times d}$, and the fixed point iteration
\begin{align*}
\bZ^{(t+1)} = \phi\left(\bM\bZ^{(t)}\bW + \wt{\bX}\right)
\end{align*}
converges to $\bZ^*$ as $t \to \infty$. Futhermore, if we assume that $\|\bZ^*\|_F \leq C_0$ for some $C_0 \geq 0$, and $\bZ^{(1)} = \mathbf{0}_{n \times d}$, then for any integer $t \geq 1$,
\begin{align*}
    \|\bZ^{(t)} - \bZ^* \|_F \leq  \kappa^{t-1} C_0.
\end{align*}
\end{theorem}

\begin{proof}

By Lemma~\ref{lem:tensor_trick}, we can deduce that
\begin{align}\label{eq:tmp_fix}
    \vect(\bM\bZ\bW) = (\bW^\top \otimes \bM) \vect(\bZ).
\end{align}
Note that we assume $\lambda_{\max}(|\bW|) < 1$, and Lemma~\ref{lem:lap_eigen} implies that $\lambda_{\max}(\bM) := \kappa < 1$. By Lemma~\ref{lem:spectrum_kronecker}, we can conclude that
\begin{align*}
    \lambda_{\max}(|\bW^\top \otimes \bM|) = &~ \lambda_{\max}(|\bW^\top|)\lambda_{\max}(|\bM|)  \\
    = &~ \lambda_{\max}(|\bW|)\lambda_{\max}(|\bM|)  \\
    = &~ \kappa < 1.
\end{align*}

Let $g(\bz) := (\bM \otimes \bW^\top)\bz + \wt{\bX}$. Note that $\phi: \R \to \R$ is $1$-Lipschitz and $g$ is an affine mapping. Since $\lambda_{\max}(|M \otimes \bW^\top|) < 1$, the mapping $g$ is contractive, i.e., for any $\bz_1, \bz_2 \in \R^{nd}$,
\begin{align}\label{eq:tmp_iter}
     \| \phi(g(\bz_1)) - \phi(g(\bz_2)) \|_2 \leq  \kappa \| \bz_1 - \bz_2 \|_2,
\end{align}
then by the well-known Banach fixed point theorem, the equation
\begin{align*}
    \vect(\bZ) = \phi \left((\bM \otimes \bW^\top)\vect(\bZ)+ \wt{\bX}\right)
\end{align*}
has a unique solution $\bZ^*$, and the fixed point iteration can converges to it. By Eq.~\eqref{eq:tmp_fix}, we prove the first result.

Note that, by tensor trick, we have $\vect(\bZ^{(t)}) = \phi(g(\vect(\bZ^{(t-1)})))$ and $\vect(\bZ^{*}) = \phi(g(\vect(\bZ^{*})))$.
Next, by Eq.~\eqref{eq:tmp_iter} and $\|\vect(\cdot)\|_2 = \|\cdot\|_F$, we have
\begin{align*}
 \|\bZ^{(t)} - \bZ^* \|_F = &~ \| \vect(\bZ^{(t)}) - \vect(\bZ^*) \|_2 \\
 = &~   \|\phi(g(\vect(\bZ^{(t-1)}))) - \phi(g(\vect(\bZ^{*})))\|_2 \tag{Simple algebra}\\
 \leq &~ \kappa \|\vect(\bZ^{(t-1)}) \vect(\bZ^*)\|_2 \tag{Contraction of $\phi(g(\cdot))$} \\
 \leq &~ \cdots \\
 \leq &~ \kappa^{t-1} \|\vect(\bZ^{(1)}) - \vect(\bZ^*)\|_2 \\
 = &~ \kappa^{t-1} \|\vect(\bZ^*)\|_2 \tag{$\bZ^{(1)} = \mathbf{0}_{n \times d}$}\\
 = &~  \kappa^{t-1} \| \bZ^* \|_F \tag{$\|\vect(\cdot)\|_2 = \|\cdot\|_F$}\\
 \leq &~ \kappa^{t-1} C_0. \tag{$\|\bZ^*\|_F \leq C_0$}\
\end{align*}
Thus we complete the proof.

\end{proof}

\section{Oversmoothing Analysis}\label{app:over_smooth}
\begin{theorem}[Sufficient condition for nonidentical node features, restatement of Theorem~\ref{thm:nonid_node_feature}]
    Let $\bM \in \R^{n \times n}$ be the normalized hypergraph Laplcaian matrix of an admissible hypergraph.  Let $\phi:\R\to\R$ be a strictly increasing nonexpansive activation function. Suppose that the weight matrix $\bW \in \R^{d \times d}$ of an IHGNN satisfies $\lambda_{\max}(|\bW|) < 1,$ then for any $\wt{\bX} \in \R^{d\times d}$ satisfying $\bx_i \neq \bx_j$ for some $i, j \in [n]$, there does not exists $\bz_0 \in \R^d$, such that $\bZ^* = \mathbf{1}_n\bz_0^\top$.
\end{theorem}

\begin{proof}
    Assume for contradiction that there exists a vector 
    $\bz_0 \in \R^{d}$ such that the constant-row matrix
    $\bZ^{*}= \mathbf{1}_n \bz_0^{\top}$ 
    is a fixed point of the implicit layer, that is
    \begin{align*}
        \bZ^{*} &= \phi\left(\bM \bZ^{*} \bW + \wt \bX\right).
    \end{align*}
    Because $\bM$ is the normalized hypergraph Laplacian matrix and it is row-stochastic, 
    $\bM\mathbf{1}_n = \mathbf{0}_n$.  Hence
    \begin{align*}
        \bM \bZ^{*} 
        &= \bM \left(\mathbf{1}_n \bz_0^{\top}\right)
        = (\bM\mathbf{1}_n) \bz_0^{\top}
        = \mathbf{0}_{n\times d}.
    \end{align*}
    Substituting this into the fixed-point equation gives, for every
    $i\in[n]$, $\bz_0 = \phi\left(\wt \bx_i\right)$,
    where $\wt \bx_i^{\top}$ is the $i$-th row of $\wt \bX$.
    Since $\phi$ is strictly increasing, it is injective, so the above
    equalities imply
    \begin{align*}
        \wt \bx_i = \wt \bx_j
        \quad \forall i,j\in[n].
    \end{align*}
    This contradicts the hypothesis that there exist indices
    $i\neq j$ with $\wt \bx_i\neq\wt \bx_j$.
    Therefore no constant-row fixed point of the form
    $\bZ^{*}=\mathbf{1}_n \bz_0^{\top}$ can exist.
\end{proof}

\begin{theorem}[Expressivity of IHGNN, restatement of Theorem~\ref{thm:expressivity}]
    Let $\bM \in \R^{n \times n}$ be the normalized hypergraph Laplcaian matrix of an admissible hypergraph. Let $K \in \N$. For every $K$-order polynomial filter function $p(\bX) := (\sum_{k=0}^K \theta_k \bM^k)\bX$ with arbitrary coefficients $\{\theta_k\}_{k=0}^K$ and input feature matrix $\bx \in \R^{n \times d}$, there exists an IHGNN with identity activation can express it.
\end{theorem}
\begin{proof}
    Fix an input feature matrix $\bX \in\R^{n \times d}$ and denote 
    $p(\bX)=(\sum_{k=1}^K \theta_k \bM^k)\bX$.
    We construct an IHGNN with identity activation that produces
    exactly this mapping. 
    First, we set the hidden state dimension $d_h := (K+1)d$, and we write the hidden state
    \begin{align*}
        \bZ = \begin{bmatrix}
            \bZ^{(0)} & \bZ^{(1)} & \cdots &\bZ^{(K)}
        \end{bmatrix}
    \end{align*}
    where each $\bZ^{(k)}$ is a block matrix of size $n \times d$. Note that here the supscript $(k)$ denotes $k$-th block matrix, not the iteration number of the fixed point iteration. However, we will show that they coincides with each other.
    We define $\bTheta_1 \in \R^{d \times d_h}$ as
    \begin{align*}
        \bTheta_1 = \begin{bmatrix}
            \bI_d & \mathbf{0}_{d \times d} & \cdots & \mathbf{0}_{d \times d}
        \end{bmatrix}
    \end{align*}
    and $\bb = \mathbf{0}_{d_h}$.
    We define the weight matrix $\bW \in \R^{d_h \times d_h}$ as
    \begin{align*}
        \bW := 
        \begin{bmatrix}
            \mathbf{0}_{d \times d} & \bI_{d} &  \mathbf{0}_{d \times d} & \mathbf{0}_{d \times d} & \cdots &  \mathbf{0}_{d \times d} \\
            \mathbf{0}_{d \times d} & \mathbf{0}_{d \times d} & \bI_{d}& \mathbf{0}_{d \times d} & \cdots &  \mathbf{0}_{d \times d} \\
            \mathbf{0}_{d \times d} & \mathbf{0}_{d \times d} & \mathbf{0}_{d \times d} & \bI_{d} & \cdots &  \mathbf{0}_{d \times d} \\
            \vdots & \vdots & \vdots &  \ddots &  \ddots &   \vdots \\
            \mathbf{0}_{d \times d} & \mathbf{0}_{d \times d} & \mathbf{0}_{d \times d} & \mathbf{0}_{d \times d} & \cdots &  \bI_{d} \\
            \mathbf{0}_{d \times d} & \mathbf{0}_{d \times d}& \mathbf{0}_{d \times d} & \mathbf{0}_{d \times d} & \cdots &  \mathbf{0}_{d \times d}
        \end{bmatrix}.
    \end{align*}

    Next, the fixed-point equilibrium equation 
    \begin{align*}
        \bZ  = &~ \phi\left(\bM \bZ \bW+ \bX\bTheta_1 + \bb \right)
    \end{align*}
    can be blockwisely written as
    \begin{align*}
        \begin{cases}
            \bZ^{(0)} = \bX, \\
            \bZ^{(1)} = \bM \bZ^{(0)} = \bM \bX, \\
            \bZ^{(2)} =  \bM \bZ^{(1)} = \bM^2\bX, \\        
            \quad \vdots \\
            \bZ^{(K)} = \bM \bZ^{(K-1)} = \bM^K\bX. \\  
        \end{cases}
    \end{align*}

    We define $\bTheta_2 \in \R^{d_h \times d}$ as
    \begin{align*}
        \bTheta_2 = \begin{bmatrix}
            \theta_0 \bI_d & \theta_1 \bI_d & \cdots & \theta_K \bI_d
        \end{bmatrix}^\top.
    \end{align*}

    Then we can conclude that
    \begin{align*}
        \bTheta_2\bZ = (\sum_{k=0}^K \theta_k \bM^k)\bX.
    \end{align*}
    Thus we complete the proof.
\end{proof}


\section{Transductive Generalization Bound}\label{app:gen_bound}

\begin{assumption}\label{as:ass_generalization:app}
     We assume the following conditions hold.
     \begin{itemize}
         \item \textbf{Bounded input features}: The input node matrix $\bX \in \R^{n \times d}$ satisfies, for each $i \in[n]$, $\|\bx_i\|_2 \leq C_X$ for some $C_X > 0$.
         \item \textbf{Bounded trainable parameters}: The trainable parameters satisfies $\|\bTheta_1\|_F \leq \rho_1, \|\bTheta_2\|_F \leq \rho_2, \|\bb\|_2 \leq C_b$ for some $\rho_1, \rho_2, C_b > 0$, and $\|\bW\| \leq \kappa$ for some $\kappa \in [0,1)$, and their dimensions satisfies $d = d_h = d'$.
         \item \textbf{Lipschitz loss}: The loss function $\ell: \R \times \R \to [0,\infty)$ is $C_\ell$-Lipschitz.
         \item \textbf{Lipschitz activation}: The activation function $\phi: \R \to \R$ is $1$-Lipschitz.
     \end{itemize}
\end{assumption}

\begin{lemma}
    Assume that Assumption~\ref{as:ass_generalization} hold. We have
    \begin{align*}
         \E_{\bVareps}\left[\langle \bVareps, \wt{\bX}\rangle \right] \leq \sqrt{n}\rho_1  C_X + \sqrt{nd}C_b. 
    \end{align*}
    where $\bVareps \in \R^{n \times d}$ is a random matrix, each entry of which is a Rademacher random variable. 
\end{lemma}
\begin{proof}
    We can show that
    \begin{align*}
        \E_{\bVareps}\left[\langle \bVareps, \wt{\bX}\rangle \right] = &~ \E_{\bVareps}\left[\langle \bVareps,  \bX\bTheta_1 + \mathbf{1}_{n} \bb^\top \rangle \right] \\
        = &~ \E_{\bVareps}\left[\langle \bVareps,  \bX\bTheta_1\rangle \right] + \E_{\bVareps}\left[\langle \bVareps, \mathbf{1}_{n} \bb^\top \rangle \right].
    \end{align*}
    
    We bound the two terms on the right hand side separately.
    \begin{align*}
        \E_{\bVareps}\left[\langle \bVareps \bX^\top,  \bTheta_1\rangle \right]  \leq &~ \E_{\bVareps}\left[\|\bVareps \bX^\top\|_F \|\bTheta_1\|_F \right] \\
        = &~   \|\bTheta_1\|_F \cdot \E_{\bVareps}\left[\|\bVareps \bX^\top\|_F \right] \\
        \leq &~  \rho_1 \cdot \E_{\bVareps}\left[\|\bVareps \bX^\top\|_F \right] \\
        \leq &~ \rho_1 \cdot \left(\E_{\bVareps} \left[\sum_{i=1}^n \|\bx_i\|_2^2\right] \right)^{1/2} \\
        \leq &~ \rho_1 \sqrt{n}C_X.
    \end{align*}

    Similary, we can show that
    \begin{align*}
        \E_{\bVareps}\left[\langle \bVareps, \mathbf{1}_{n} \bb^\top \rangle \right] = &~ \E_{\bVareps}\left[\langle  \bb, \bVareps^\top \mathbf{1}_{n} \rangle \right] \\
        \leq &~\|\bb\|_2 \E_{\bVareps}\left[\|\bVareps^\top \mathbf{1}_{n} \|_2 \right] \\
        \leq &~ C_b \E_{\bVareps}\left[\|\bVareps^\top \mathbf{1}_{n} \|_2 \right] \\
        \leq &~ C_b \left(\E_{\bVareps} \left[\sum_{i=1}^n \sum_{j=1}^d |\bVareps_{i,j}|^2\right]\right)^{1/2} \\
        = &~ C_b \cdot \sqrt{nd}.
    \end{align*}
    Hence we complete the proof.
\end{proof}

\begin{lemma}[Rademacher complexity of the implicit layer]\label{lem:rad_imp:app}
    We define
    \begin{align*}
        \cF_T := \{ f(\cdot): f(\wt{\bX}) = \bZ^{(t+1)} := \phi(\bM \bZ^{(t)} \bW +  \wt{\bX}), t \in [T] \}.
    \end{align*}
    Assume that Assumption~\ref{as:ass_generalization} hold. Then we have
    \begin{align*}
        \cR_n^\mathrm{coord}(\cF_T) \leq &~ \frac{\rho_1 C_x + \sqrt{d}C_b}{\sqrt{n}(1-\kappa)}.
    \end{align*}
\end{lemma}
\begin{proof}
    We can show that
    \begin{align*}
        n \cR_n^\mathrm{coord}(\cF_T) =  &~ \E_{\bVareps} \left[\sup_{f \in \cF_T} \langle \bVareps, f(\wt{\bX})\rangle \right] \\
        =  &~ \E_{\bVareps} \left[\sup_{f \in \cF_{T-1}} \langle \bVareps, \bM f(\wt{\bX})\bW + \wt{\bX}\rangle \right] \\
        =  &~ \E_{\bVareps} \left[\sup_{f \in \cF_{T-1}} \langle \bVareps, \bM f(\wt{\bX})\bW \rangle\right] +  \E_{\bVareps} \left[\langle \bVareps, \wt{\bX}\rangle \right] \\
        =  &~ \E_{\bVareps} \left[\sup_{f \in \cF_{T-2}} \langle \bVareps, \bM (\bM f(\wt{\bX})\bW + \wt{\bX})\bW \rangle\right] +  \E_{\bVareps} \left[\langle \bVareps, \wt{\bX}\rangle \right]  \\
        =  &~ \E_{\bVareps} \left[\sup_{f \in \cF_{T-2}} \langle \bVareps, \bM^2 f(\wt{\bX})\bW^2 \rangle\right] + \E_{\bVareps} \left[\langle \bVareps, \bM\wt{\bX}\bW\rangle \right]+ \E_{\bVareps} \left[\langle \bVareps, \wt{\bX}\rangle \right] \\
        = &~ \cdots \\
         =  &~ \E_{\bVareps} \left[ \langle \bVareps, \bM^{T-1} f(\wt{\bX})\bW^{T-1} \rangle\right] + \cdots + \E_{\bVareps} \left[\langle \bVareps, \bM\wt{\bX}\bW\rangle \right]+ \E_{\bVareps} \left[\langle \bVareps, \wt{\bX}\rangle \right] \\
        = &~ \sum_{t=1}^{T} \E_{\bVareps} \left[\langle \bVareps, \bM^{t-1} \wt{\bX}\bW^{t-1} \rangle \right] \\
        \leq &~ \sum_{t=1}^{T} \kappa^{t-1} \E_{\bVareps} \left[\langle \bVareps, \wt{\bX}\rangle \right] \\
        = &~ \frac{1-\kappa^T}{1-\kappa}(\sqrt{n}\rho_1 C_x + \sqrt{nd}C_b) \\
        \leq &~ \frac{1}{1-\kappa}(\sqrt{n}\rho_1 C_x + \sqrt{nd}C_b).
    \end{align*}
    Dividing $n$ on both sides gives
    \begin{align*}
        \cR_n^\mathrm{coord}(\cF_T) \leq \frac{\rho_1 C_x + \sqrt{d}C_b}{\sqrt{n}(1-\kappa)}.
    \end{align*}
    Thus the proof is complete.
\end{proof}

\begin{theorem}[Transductive generalization bound of IHGNN, restatement of Theorem~\ref{thm:gen_bound_IHGNN}]\label{thm:gen_bound_IHGNN:app}
    We assume that the hypergraph is admissible and Assumption~\ref{as:ass_generalization:app} are satisfied. Let $\cH$ be the hypothesis class of IHGNN models defined on the given hypergraph. Let $c_0 := \sqrt{\frac{32\log(4e)}{3}} < 5.05$.
     Let $P := \frac{1}{s} + \frac{1}{u}$, $Q := \frac{s+u}{(s+u-1/2)(1-1/(2\max\{s,u\}))}.$ Then, for any $\delta > 0$, with
    probability at least $1-\delta$ over the choice of the training set $\{\bx_i\}_{i=1}^{s+u} \cup \{y_i\}_{i=1}^{u}$, for all $f \in \cH$, we have
    \begin{align*}
    \cL_u(f) \leq &~ \wh{\cL}_s(f) + \frac{\sqrt{2}\rho_2 C_\ell(\rho_1 C_x + \sqrt{d}C_b)}{(1-\kappa)\sqrt{s+u}} + c_0 P \sqrt{\min\{s,u\}}+ \sqrt{\frac{PQ}{2}\log\frac{1}{\delta}}.
    \end{align*}
\end{theorem}

\begin{proof}
    By Lemma~\ref{lem:rad_imp:app}, Lemma~\ref{lem:contra_vect:app}, Lemma~\ref{lem:trans_gen_b:app}, and Lemma~\ref{def:rad_ineq:app}, we can show that
    \begin{align*}
        \cL_s(f) \leq &~ \wt{\cL}_u(f) + \wt{R}_{s+u}(\ell \circ \cH) + c_0 P \sqrt{\min\{s, u\}} + \sqrt{\frac{PQ}{2} \log \frac{1}{\delta}} \\
        \leq &~ \wt{\cL}_u(f) + R_{s+u}(\ell \circ \cH) + c_0 P \sqrt{\min\{s, u\}} + \sqrt{\frac{PQ}{2} \log \frac{1}{\delta}} \\
        = &~ \wt{\cL}_u(f) + R_{s+u}(\ell \circ \phi \circ \cF_T) + c_0 P \sqrt{\min\{s, u\}} + \sqrt{\frac{PQ}{2} \log \frac{1}{\delta}} \\
        \leq &~ \wt{\cL}_u(f) + \sqrt{2} \rho_2 C_\ell \cdot R_{s+u}^\mathrm{coord}(\cF_T) + c_0 P \sqrt{\min\{s, u\}} + \sqrt{\frac{PQ}{2} \log \frac{1}{\delta}} \\
        \leq &~ \wt{\cL}_u(f) +  \frac{\sqrt{2} \rho_2 C_\ell(\rho_1 C_x + \sqrt{d}C_b)}{\sqrt{s+u}(1-\kappa)} + c_0 P \sqrt{\min\{s, u\}} + \sqrt{\frac{PQ}{2} \log \frac{1}{\delta}}.
    \end{align*}
    Thus we complete the proof.
\end{proof}

\begin{corollary}[Asymptotic transductive generalization bound of IHGNN, restatement of Corollary~\ref{cor:asy_gen_bound_IHGNN}]\label{cor:asy_gen_bound_IHGNN:app}
    Under the same conditions in Theorem~\ref{thm:gen_bound_IHGNN:app}.
    For sufficiently large training-set size $s$ and testing-set size $u$, for any $\delta > 0$, with
    probability at least $1-\delta$ over the choice of the training set, for all $f \in \cH$, we have
    \begin{align*}
    \cL_u(f) \leq &~ \wh{\cL}_s(f) + O\left(\frac{d}{s+u}\right)^{\frac{1}{2}} + O \left(\frac{\log(1/\delta)}{\min\{s,u\}}\right)^{\frac{1}{2}}.
    \end{align*}
\end{corollary}

\begin{proof}
    It is not hard to see when $s$ and $u$ are sufficiently large, we have $Q = O(1)$.
    
    First, it is not hard to see that
    \begin{align*}
        \frac{\sqrt{2}\rho_2 C_\ell(\rho_1 C_x + \sqrt{d}C_b)}{(1-\kappa)\sqrt{s+u}} = O\left(\sqrt{\frac{d}{s+u}}\right).
    \end{align*}

    Next, we can show that
    \begin{align*}
        c_0 P \sqrt{\min\{s,u\}}+ \sqrt{\frac{PQ}{2}\log\frac{1}{\delta}} = &~ c_0 \left(\frac{1}{s}+\frac{1}{u}\right)\sqrt{\min\{s,u\}} + O\left(\sqrt{\left(\frac{1}{s}+\frac{1}{u}\right)\log\frac{1}{\delta}}\right) \\
        = &~ O\left(\frac{1}{\sqrt{s}}+\frac{1}{\sqrt{u}}\right) + O\left(\sqrt{\left(\frac{1}{s}+\frac{1}{u}\right)\log\frac{1}{\delta}}\right) \\
        = &~ O\left(\sqrt{\left(\frac{1}{s}+\frac{1}{u}\right)\log\frac{1}{\delta}}\right).
    \end{align*}
    
    Thus, by Theorem~\ref{thm:gen_bound_IHGNN:app}, we complete the proof.
\end{proof}

\section{Training of IHGNN}\label{app:training}
\begin{theorem}[Scaled well-posedness of IHGNN, restatement of Theorem~\ref{thm:scale_wellpose}]
Suppose that the activation function $\phi: \R \to \R$ is positively homogeneous and nonexpansive. If an IHGNN model with weights $\bW, \bTheta_1, \bTheta_2, \bb$ satisfies $ \lambda_{\max}(|\bW|) < 1 $, then there exists an equivalent IHGNN model with weights $\wt{\bW}, \wt{\bTheta}_1, \wt{\bTheta_2}, \wt{\bb}$ such that $ \|\wt{\bW}\|_{\infty} < 1$, 
and both models produce identical outputs for all same inputs.
\end{theorem}
\begin{proof}
Let $\alpha \in (0,1)$ be a scaling factor such that $\tilde{\bW} := \alpha \bW$ satisfies
\begin{align*}
    \|\tilde{\bW}\|_{\infty} = \alpha \|\bW\|_{\infty} < \alpha.
\end{align*}

We define $\wt{\bTheta_1} := \alpha \bTheta_1$, and $\wt{\bb} := \alpha \bb$. Then it is not hard to see that for any $\bX \in \R^{n \times d}$, the unique solution $\wt{\bZ}^*$ of the equation
\begin{align*}
    \wt{\bZ} = \phi\left(\bM \wt{\bZ} \wt{\bW} + \bX \wt{\bTheta}_1 +\wt{\bb}\right)
\end{align*}
satisfies $\wt{\bZ}^* = \alpha \bZ^*$, where $\bZ^*$ is the fixed point solution of
$
    {\bZ} = \phi\left(\bM {\bZ} {\bW} + \bX {\bTheta}_1 +{\bb}\right).
$
Hence, by defining $\wt{\bTheta}_2 := \alpha^{-1} \bTheta_2$ , we complete the proof.
\end{proof}

\section{Heatmap Comparisons of different learning rates and dropout settings}

\begin{figure*}[!ht]
    \centering
    \begin{subfigure}[t]{0.89\textwidth}
        \centering
        \includegraphics[width=\textwidth]{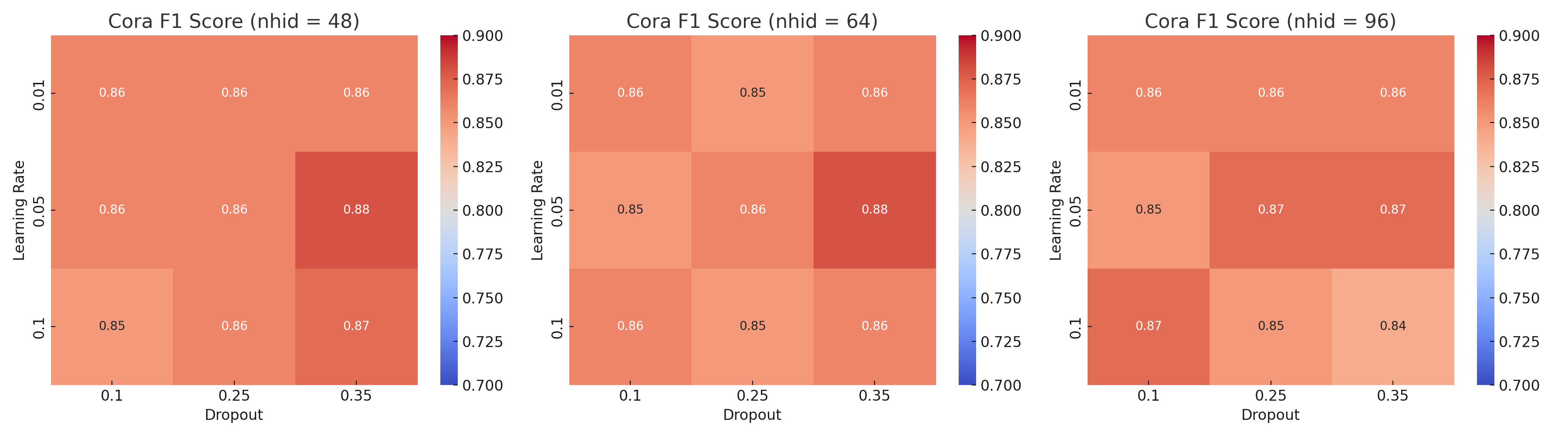}
    \end{subfigure}
    \hfill
    \begin{subfigure}[t]{0.89\textwidth}
        \centering
        \includegraphics[width=\textwidth]{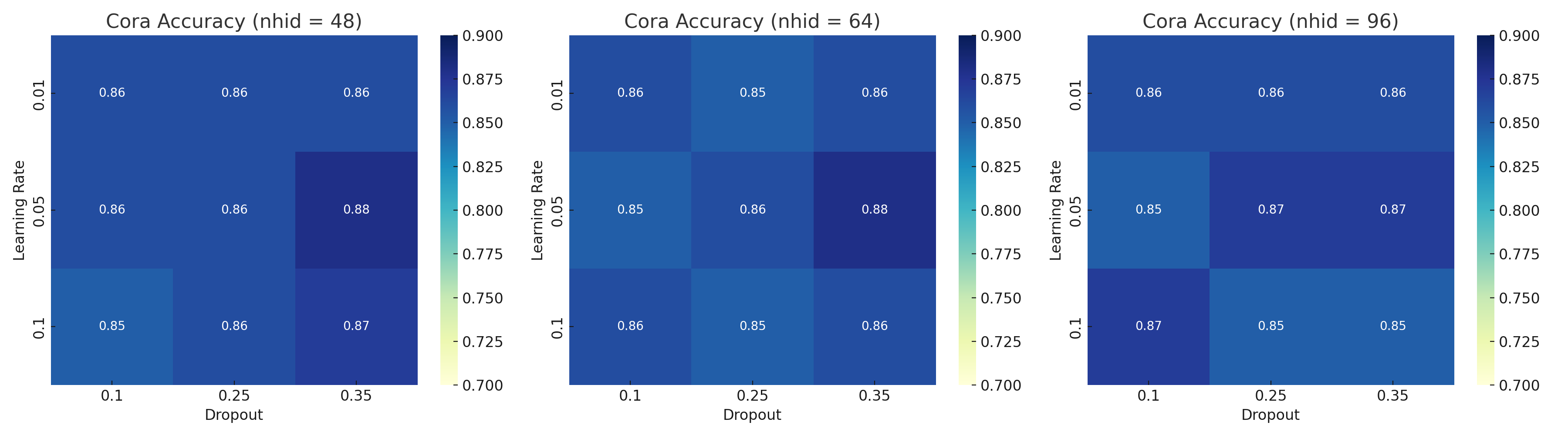}
    \end{subfigure}
    \caption{Heatmaps of \textit{F1 Score} (top) and \textit{Accuracy} (bottom) across different learning rates and dropout settings for IHGNN on \textit{Cora}.}
    \label{fig:cora}
\end{figure*}

\begin{figure*}[!ht]
    \centering
    \begin{subfigure}[t]{0.89\textwidth}
        \centering
        \includegraphics[width=\textwidth]{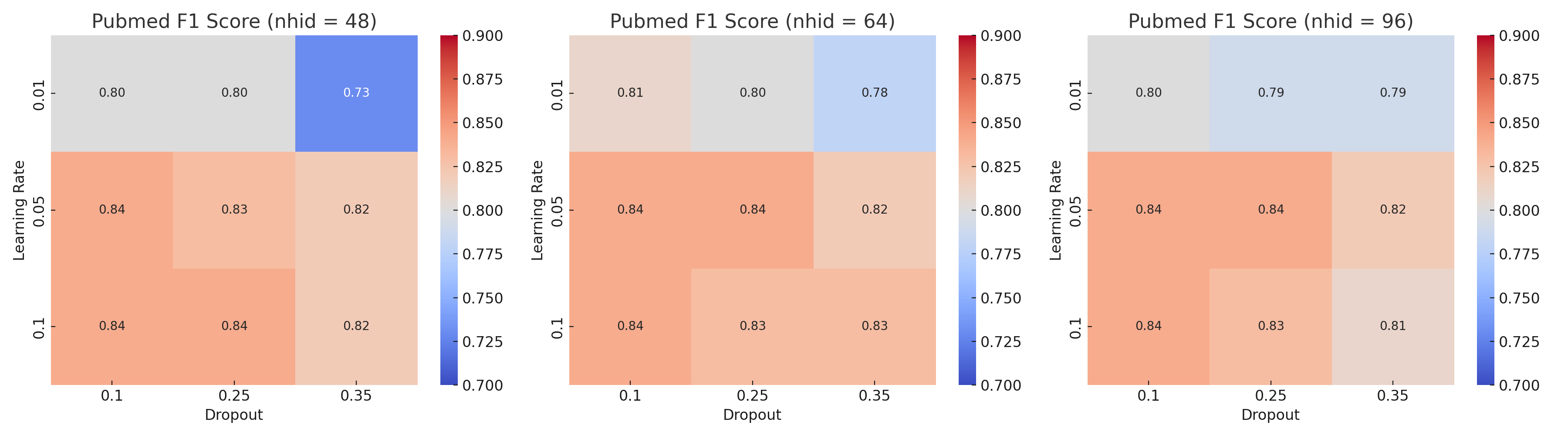}
    \end{subfigure}
    \hfill
    \begin{subfigure}[t]{0.89\textwidth}
        \centering
        \includegraphics[width=\textwidth]{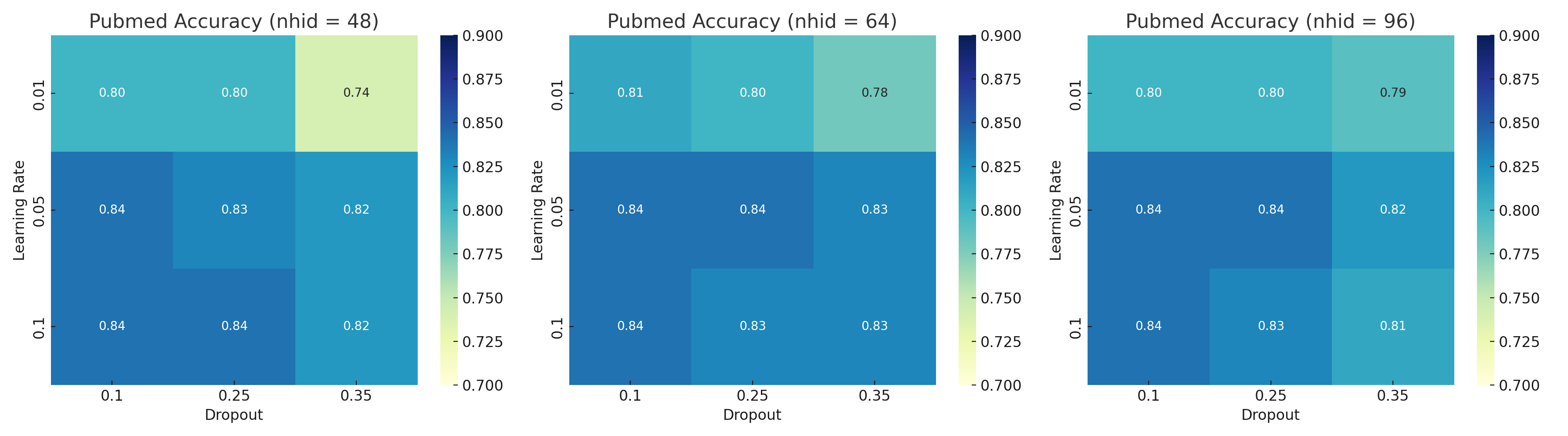}
    \end{subfigure}
    \caption{Heatmaps of \textit{F1 Score} (top) and \textit{Accuracy} (bottom) across different learning rates and dropout settings for IHGNN on \textit{Pubmed}.}
    \label{fig:pubmed}
\end{figure*}

\begin{figure*}[!ht]
    \centering
    \begin{subfigure}[t]{0.89\textwidth}
        \centering
        \includegraphics[width=\textwidth]{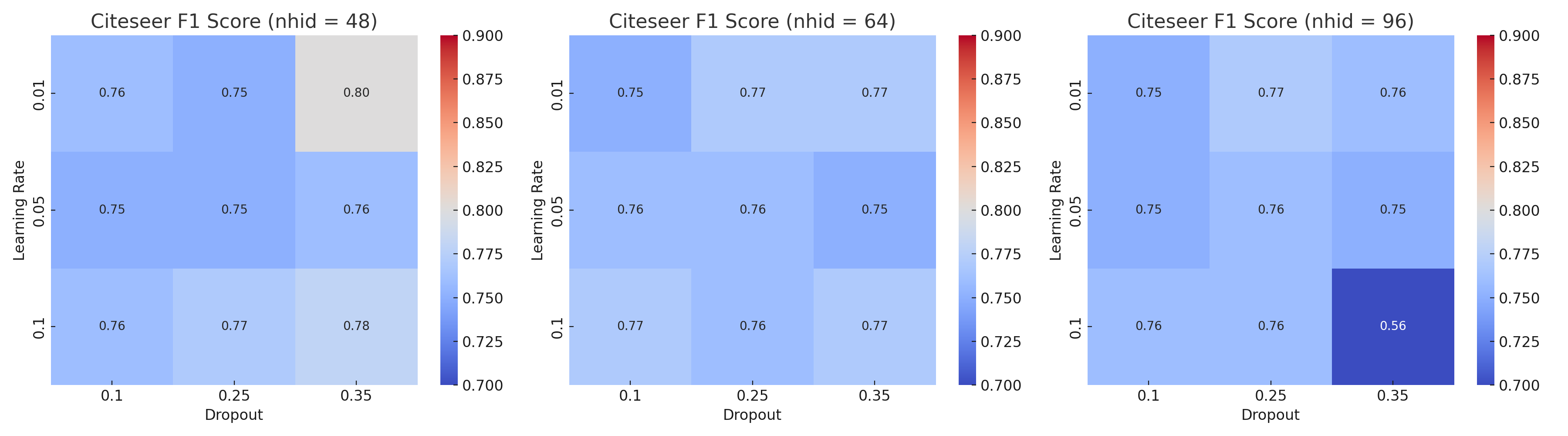}
    \end{subfigure}
    \hfill
    \begin{subfigure}[t]{0.89\textwidth}
        \centering
        \includegraphics[width=\textwidth]{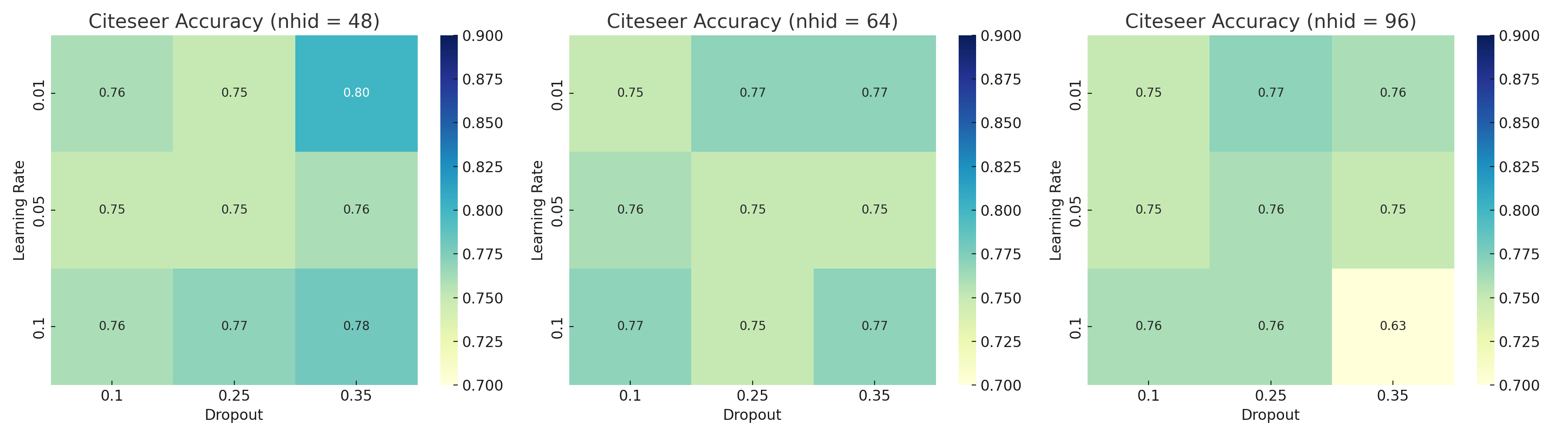}
    \end{subfigure}
    \caption{Heatmaps of \textit{F1 Score} (top) and \textit{Accuracy} (bottom) across different learning rates and dropout settings for IHGNN on \textit{Citeseer}.}
    \label{fig:citeseer}
\end{figure*}

\end{document}